\pdfoutput=1

\documentclass[11pt]{article}

\usepackage[final]{acl}
\usepackage{times}
\usepackage{adjustbox}
\usepackage{pdfpages}
\usepackage{afterpage}
\usepackage{colortbl}
\usepackage{latexsym}
\usepackage[most]{tcolorbox}
\usepackage{multirow}
\usepackage{mdframed}
\usepackage{xspace}
\usepackage{algorithmic}
\usepackage{algorithm}
\definecolor{promptcolor}{RGB}{240, 240, 240} 
\usepackage{xcolor}
\usepackage[T1]{fontenc}

\usepackage{pifont} 
\usepackage{enumitem}
\newcommand{\cmark}{\textcolor{green}{\ding{51}}}  
\newcommand{\xmark}{\textcolor{red}{\ding{55}}}    

\usepackage[utf8]{inputenc}

\usepackage{amsmath}     
\usepackage{array}       
\usepackage{tabularx}    
\usepackage{booktabs}    
\usepackage{multirow}    
\usepackage{colortbl}    
\usepackage{xcolor}      
\usepackage{caption}     

\usepackage{makecell}

\usepackage{footmisc}

\usepackage{microtype}

\usepackage[normalem]{ulem} 

\usepackage{inconsolata}

\usepackage{graphicx}
\usepackage{subcaption}  

\title{%
  \raisebox{-0.3\height}{\includegraphics[height=1.7em]{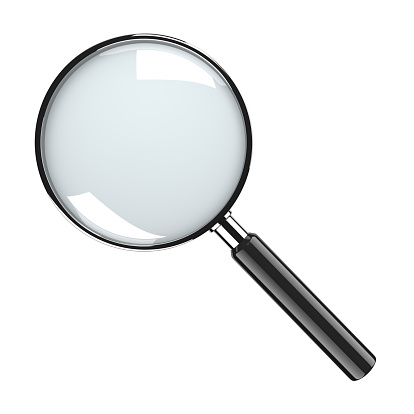}}\,Finding Needles in Images: Can Multimodal LLMs Locate Fine Details?%
}

\usepackage{hyperref}  

\author{
  \textbf{Parth Thakkar\textsuperscript{1}}\thanks{Equal contribution as co-first authors.} \quad
  \textbf{Ankush Agarwal\textsuperscript{1}}\footnotemark[1] \quad
  \\
  \textbf{Prasad Kasu\textsuperscript{1}} \quad
  \textbf{Pulkit Bansal\textsuperscript{1,2}}\thanks{Work done during internship at Fujitsu Research India} \quad
  \textbf{Chaitanya Devaguptapu\textsuperscript{1}} \\
  \textsuperscript{1}Fujitsu Research India \quad
  \textsuperscript{2}Indian Institute of Technology Patna \\
  \href{https://ast-fri.github.io/needles-in-images/}{\textit{Finding Needles in Images - Blog Post}} \\
  \texttt{\{parth.thakkar, ankush.agarwal\}@fujitsu.com}
}

\newcommand{ \datasetname }{NiM-Benchmark\xspace}
\newcommand{ \methodname }{Spot-IT\xspace}

\begin{document}
\maketitle

\renewcommand{\thefootnote}{}
\footnotetext{Code: \href{https://github.com/ast-fri/needles-in-images}{GitHub Repo}, Data: \href{https://huggingface.co/datasets/AST-FRI/needles-in-images}{Hugging Face Dataset}}
\renewcommand{\thefootnote}{\arabic{footnote}}

\begin{abstract}
While Multi-modal Large Language Models (MLLMs) have shown impressive capabilities in document understanding tasks, their ability to locate and reason about fine-grained details within complex documents remains understudied. Consider searching a restaurant menu for a specific nutritional detail or identifying a disclaimer in a lengthy newspaper article — tasks that demand careful attention to small but significant details within a broader narrative, akin to Finding Needles in Images (NiM). To address this gap, we introduce \datasetname, a carefully curated benchmark spanning diverse real-world documents including newspapers, menus, and lecture images, specifically designed to evaluate MLLMs' capability in these intricate tasks. Building on this, we further propose \methodname, a simple yet effective approach that enhances MLLMs capability through intelligent patch selection and Gaussian attention, motivated from how humans zoom and focus when searching documents. Our extensive experiments reveal both the capabilities and limitations of current MLLMs in handling fine-grained document understanding tasks, while demonstrating the effectiveness of our approach. \methodname achieves significant improvements over baseline methods, particularly in scenarios requiring precise detail extraction from complex layouts. 
\end{abstract}

\section{Introduction}

\begin{figure}[t]
    \centering
    \includegraphics[width=0.47\textwidth]{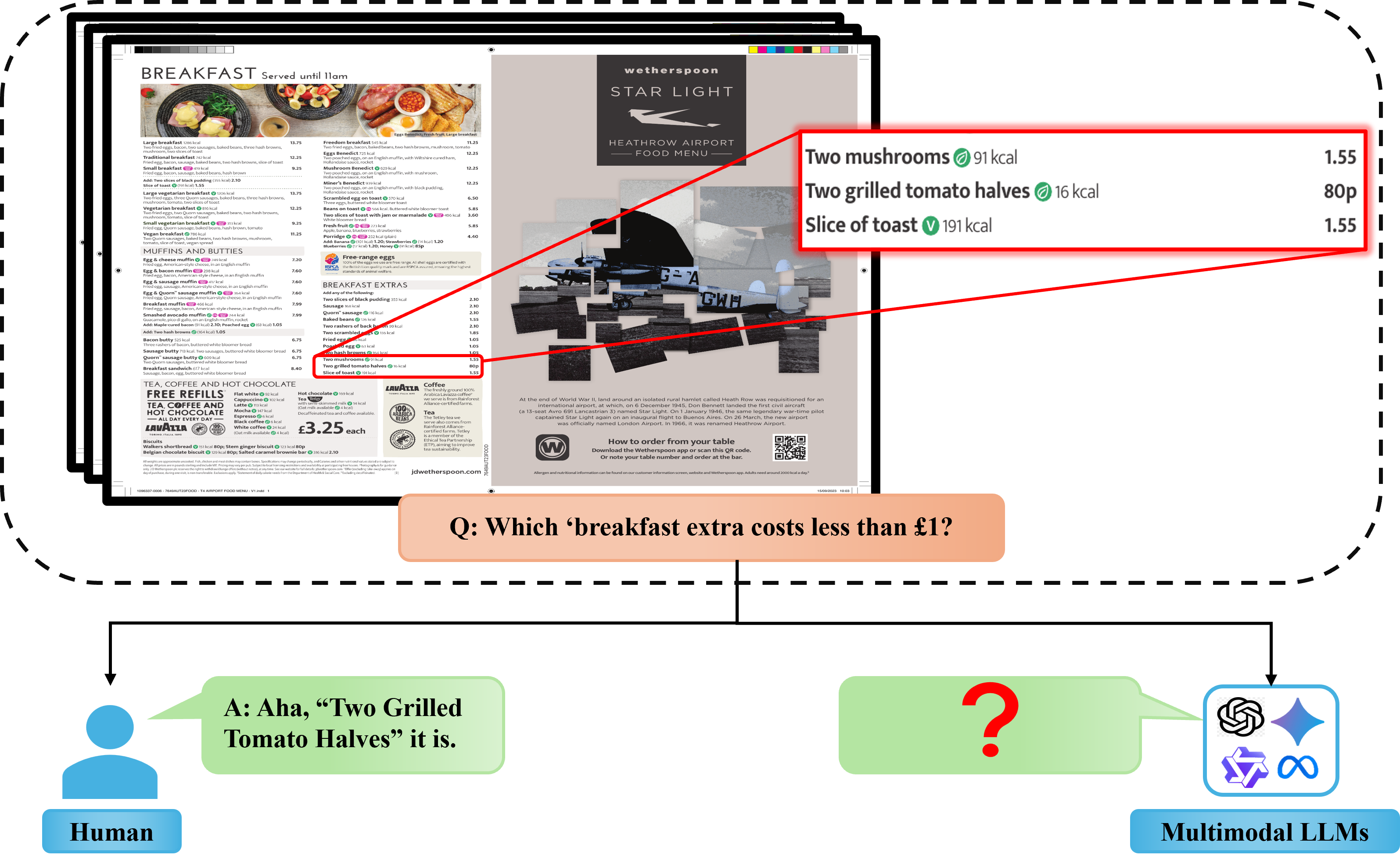}
        \caption{\footnotesize
An example of a "Needle in Images" task: finding a specific breakfast extra under \pounds1 in a restaurant menu requires precise attention to a small region while processing the entire layout. \textbf{How do MLLMs compare to humans on such tasks? We present a benchmark and a baseline method to study this.}}
    \label{fig:Intro_fig}
\end{figure}
Recent breakthroughs in Multi-modal Large Language Models (MLLMs) \cite{team2023gemini, driess2023palm, peng2023kosmos, openai2023gpt} have fundamentally transformed how machines understand and reason about visual information. These models demonstrate remarkable capabilities in visual dialogue, scene comprehension, and answering nuanced questions about visual content. For the task of Document Visual Question Answering (DocVQA)~\cite{mathew2021docvqa}, MLLMs have emerged as particularly powerful tools, interpreting visually rich documents in ways that transcend traditional text extraction methods \cite{PyPDF2, pdfminer}, enabling question answering (QA) even in documents with complex layouts and mixed text-visual elements.

\noindent While MLLMs excel at broad document comprehension, their ability to handle precise, localized information within complex documents remains an open question. Consider a seemingly simple task: \textsl{Searching a Restaurant Menu to find a breakfast extra that costs less than \pounds1} (as shown in Figure \ref{fig:Intro_fig}). This information occupies just a tiny fraction of the document's spatial extent, yet humans can efficiently locate it by combining broad visual scanning with focused attention -- quickly zeroing in on "Two Grilled Tomato Halves" as the answer. This everyday scenario highlights a fundamental challenge in document understanding: the ability to locate and reason about minute details within larger document. 

\noindent Traditional approaches based on OCR and text-extraction~\cite{smith2007overview, memon2020handwritten,pdfminer}  inherently struggle with this challenge, as they often lose the crucial connection between local details and global document structure. Even for MLLMs, despite their broad training on web-scale data~\cite{gadre2024datacomp}, processing fine-grained details within visually rich documents presents a unique challenge, especially in domain-specific documents with complex visual layout (shown in Figure \ref{fig:method-highlevel}). This difficulty stems from a fundamental tension: models must simultaneously maintain document-level context while precisely attending to minute details -- a capability that humans possess naturally but remains elusive for automated systems.

\noindent The current landscape of DocVQA research has not adequately addressed this challenge. While pioneering work like DocVQA~\cite{mathew2021docvqa} established foundations for document understanding using MLLMs, it primarily focuses on general comprehension tasks in industrial documents. Subsequent benchmarks such as SlideVQA~\cite{tanaka2023slidevqa} and MMLongBench~\cite{mammlongbench} have expanded the scope to multi-page scenarios and long-form documents, respectively. However, these benchmarks evaluate broad document comprehension rather than the specific challenge of locating and reasoning about minute details within complex layouts. This gap is particularly significant as real-world document interaction often depends on precisely locating and interpreting small but critical pieces of information within a larger context.

\noindent To address this gap, we introduce the Needles in Images Benchmark, \datasetname. This carefully curated benchmark specifically evaluates fine-grained visual reasoning in DocVQA across diverse real-world scenarios - from dense newspaper layouts to intricate restaurant menus, magazine spreads, and classroom lecture snapshots. Each document type presents unique challenges in locating and reasoning about minute details within complex visual contexts. The benchmark includes targeted question types that probe a model's capability to combine broad document understanding with precise attention to relevant local details, closely mirroring real-world information seeking scenarios.

\noindent To complement our benchmark, we propose \methodname, a simple yet effective approach that draws inspiration from human visual search behavior. Our method enhances MLLMs' ability to focus on specific document regions through a novel question-guided attention mechanism. For each input document, \methodname segments the image into patches, identifies the most relevant regions based on the query, and dynamically generates a Gaussian patch with a variable $\sigma$, adjusted using cosine similarity (as illustrated in Figure~\ref{fig:method-overview}).
 This approach enables models to better handle the dual challenges of maintaining global context while attending to local details. Below, we summarize the key contributions of our work:

\begin{enumerate}[leftmargin=*,nosep]
    \item We formalize the \textsl{Needle in an Image} challenge in DocVQA, focusing on evaluating MLLMs' ability to locate and reason about fine-grained details within complex documents. 

    \item We introduce \datasetname, a carefully curated benchmark comprising $2,970$ images and $1,180$ question-answer pairs across diverse document types including academic papers, newspapers, menu and images from classroom lectures.  Each question is specifically designed to test MLLMs' capability to extract precise details within rich visual contexts, with rigorous quality validation through both human experts and automated verification.
    

    \item We propose \methodname, a simple yet effective approach that enhances MLLMs' fine-grained reasoning capabilities through question-guided dynamic attention. Our method achieves this without requiring architectural changes to existing MLLMs, making it broadly applicable across different model architectures.

    \item Through comprehensive experiments, we demonstrate that \methodname significantly improves state-of-the-art on fine-grained detail extraction, achieving a 15.5\% improvement over GPT-4o on ArxiVQA and 21.05\% improvement on our \datasetname. These results establish new baselines for precise information extraction in DocVQA
.
\end{enumerate}

\section{Background and Related Work}\label{sec:related-work}

\noindent \textbf{Document Understanding Evolution:}
Document understanding has evolved from rule-based OCR systems \cite{smith2007overview, subramani2020survey} to sophisticated Multi-modal Large Language Models \cite{team2023gemini, openai2023gpt}. Early DocVQA datasets \cite{mathew2021docvqa, du2022calm} focused on basic text extraction and comprehension tasks, while recent benchmarks like SlideVQA \cite{tanaka2023slidevqa} and MMLongBench \cite{mammlongbench} have expanded to multi-page scenarios and long-form documents. However, these datasets primarily evaluate broad document comprehension rather than fine-grained detail extraction, which is the primary motivation for creating our benchmark. We compare our benchmark with existing ones in Table ~\ref{tab:dataset_comparison} (in Appendix).

\noindent \textbf{Fine-grained Visual Analysis in Documents:} While fine-grained visual analysis has been extensively studied in natural images \cite{yang2023finegrainedvisualprompting}, its application to document understanding remains limited. Recent visual prompting techniques \cite{wu2024visualpromptingmultimodallarge} have shown promise in directing model attention to specific image regions through bounding boxes \cite{lin2024drawandunderstandleveragingvisualprompts} or markers \cite{shtedritski2023doesclipknowred}. However, documents present unique challenges due to their hierarchical structure and complex layouts, making direct adaptation of these techniques insufficient. Our work bridges this gap by introducing both a benchmark and method specifically designed for evaluating fine-grained document analysis capabilities of MLLMs.

\noindent \textbf{Methods for Document VQA:} Current approaches to DocVQA either rely on traditional OCR-based pipelines \cite{xu2020layoutlm, huang2022layoutlmv3} or leverage end-to-end MLLMs \cite{zhang2024cream, zhang2024cfret}. For larger documents, retrieval-augmented generation (RAG) methods \cite{faysse2024colpaliefficientdocumentretrieval} have emerged as a promising direction. However, these methods typically process entire document regions without considering the granularity of relevant information, leading to inefficiencies when only small portions contain the answer. Our \methodname addresses this limitation through a question-guided attention mechanism that selectively focuses on relevant document regions.
\noindent For an extended discussion of related work, please refer to Appendix~\ref{sec:extended_related_work}.

\section{Dataset: Needle in an Image Benchmark}

\noindent
Our benchmark, \datasetname, is designed to evaluate MLLMs' ability to locate and reason about fine-grained details within complex documents. We define fine-granularity using the following ratio:

\vspace{-1em}
\begin{multline}
\text{Fine-Granularity} = \frac{\text{Area of Relevant Region}}{\text{Total Image Page Area}} \\
< 0.05
\end{multline}

\noindent
\textit{i.e.}, a task is considered fine-grained when the relevant region occupies less than 5\% of the total image area.

\noindent
In this section, we describe the dataset construction process, its characteristics, and provide an in-depth analysis.

\subsection{Dataset Construction}

\noindent Our dataset spans multiple domains including academic papers, newspapers, magazines, lecture materials, and restaurant menus, each presenting unique challenges in locating fine-grained information.

\noindent \textbf{Document Collection and Processing:} We curated documents from six diverse domains: (1) Restaurant menus with complex layouts and pricing information, (2) Recent academic papers from arXiv (2024-2025), (3) Magazines covering diverse domains with mixed text-visual content, (4) Contemporary English e-newspapers, (5) Website screenshots from the CoVA dataset \cite{kumar-etal-2022-cova}, and (6) Classroom lecture screenshots from open educational resources. Details of the domain sources are present in Table \ref{tab:domain-sources} (in Appendix).

\noindent  To ensure consistency, all documents were converted to a uniform image format while preserving visual complexity and layout using a Python library~\cite{pdf2image}. Documents example images are shown in Table \ref{tab:vqa_examples} in Appendix.

\noindent \textbf{Question-Answer Pair Generation:} We employed a hybrid approach to create high-quality question-answer pairs that specifically target fine-grained information: (1) We divided each document into variable-sized patches (2×2 to 6×6 grids) and used a MLLM with carefully crafted prompts to generate initial QA pairs focusing on localized information within each patch (2) The initial pool of QA pairs are verified by a human annotator and the irrelevant pairs were discarded. For certain domains, automated generation with filtering proved insufficient, so a team of four annotators created fine-grained questions for those domains. (3) All QA pairs underwent verification by three independent annotators to ensure accuracy, relevance, and consistency with our focus on fine-grained detail extraction.
All prompts used for dataset construction are detailed in Section \ref{sec:prompts used} in the Appendix.

\vspace{-0.2cm}
\subsection{Dataset Characteristics and Analysis}

\noindent Our dataset includes 284 documents across six domains, containing 1,180 question-answer pairs. An overview is provided in Table \ref{tab:dataset-stats}.
Each domain presents unique challenges for fine-grained information extraction, from dense multi-column newspaper layouts to technical diagrams in academic papers. 

\noindent \textbf{Question Types and Distribution:} We categorize questions into several types to assess fine-grained understanding: (1) \textsl{Inline}: Direct extraction of specific details, (2) \textsl{Boolean}: Yes/no questions about specific details, (3) \textsl{Comparative}: Comparison between nearby elements, (4) \textsl{Complex Reasoning}: Multi-step inference about document details, (5) \textsl{Commonsense}: Requiring world knowledge, and (6) \textsl{Unanswerable}: Context needed to answer is absent. Table \ref{tab:domain_distribution} in Appendix presents the distribution of question categories across domains.

\subsection{Quality Analysis}

\noindent To validate the quality of our automatically generated question-answer pairs, we conducted rigorous evaluations using two carefully curated test sets: (1) Set X containing 200 human-generated questions from existing datasets, and (2) Set Y comprising 200 samples from our dataset with balanced representation across domains (30-35 questions per domain). Our analysis encompassed three complementary dimensions:

\noindent \textbf{Response Time Analysis:} We measured response times and accuracy (EM and F1 scores) across three MLLMs (GPT-4o, Gemini-1.5-Flash, GPT-4o-mini) and human experts on Set Y. This analysis, visualized in Figure \ref{fig:time_chart_nid}, demonstrates that although human accuracy is moderately high on our dataset, it comes at the cost of increased response time.

\noindent \textbf{Question Quality Assessment:} We conducted a blind Turing test where two independent researchers evaluated a mixed set of human and machine-generated questions (Sets X and Y combined). The inter-annotator agreement (Cohen's $k$ \cite{Cohen1960} = 0.234) indicates that our generated questions are comparable to human-crafted ones in terms of quality and naturalness.

\noindent \textbf{Automated Verification:} To ensure scalable quality assessment, we employed Claude-3.5-Sonnet and Gemini-2.0-Flash as independent judges, achieving strong inter-model agreement ($k$ = 0.339). These models were specifically chosen to avoid potential biases, as they were not involved in the question generation process.

\section{Methodology: \methodname}
Finding a "needle" of information in a complex document requires a delicate balance between broad context awareness and precise attention to detail. Our method, \methodname, draws inspiration from how humans efficiently locate specific details in documents: first identifying potentially relevant regions based on the query, then focusing attention on those regions while maintaining awareness of the surrounding context. This two-stage approach enables effective extraction of fine-grained information while preserving the document's structural context.

\begin{figure*}[t]
    \centering
    \includegraphics[width=\textwidth,height=8.5
    cm]{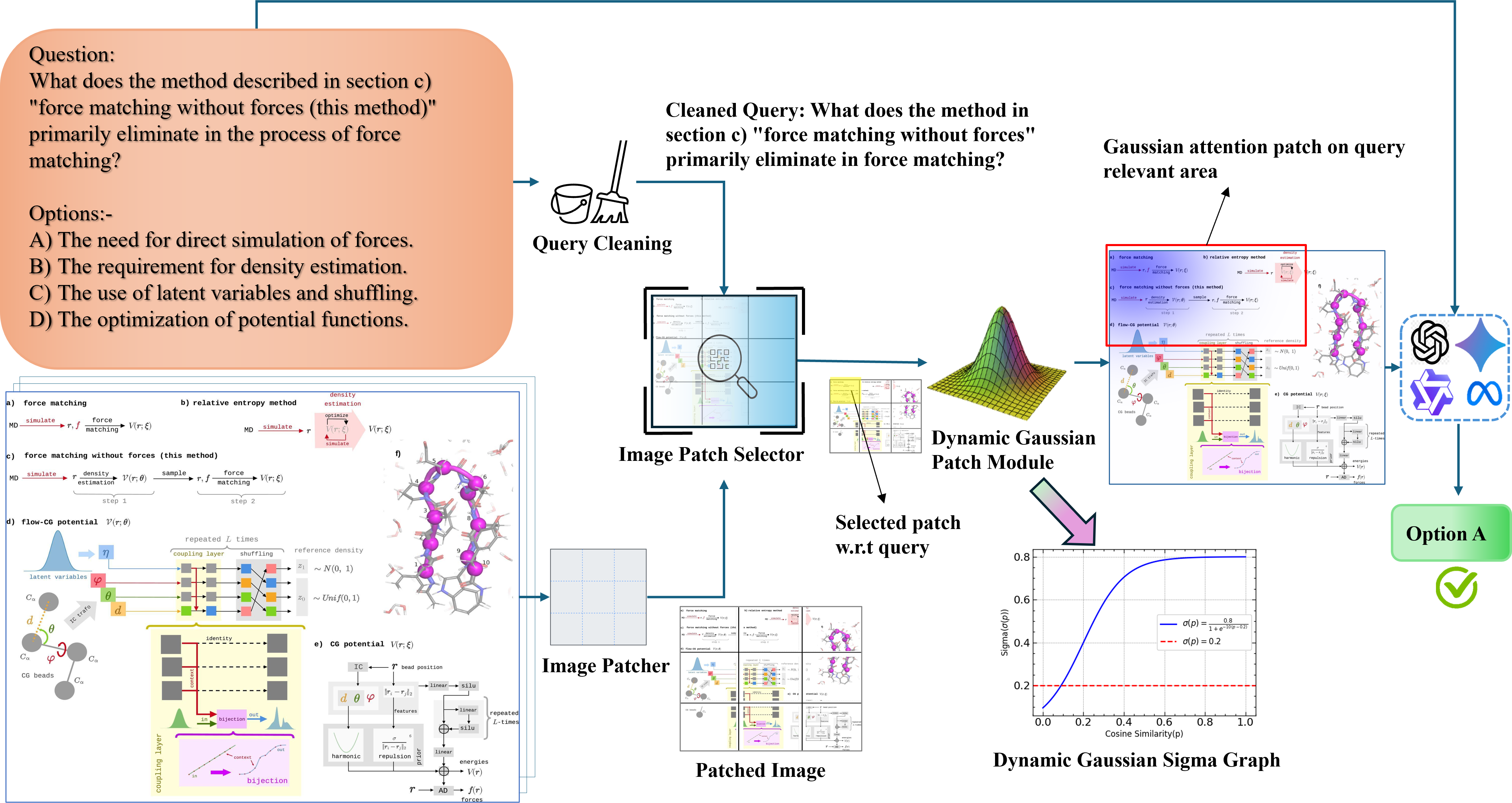}
    \caption{\footnotesize Overview of \methodname: Given a document and query, our method (1) cleans the query, (2) identifies the most relevant image patch, (3) applies an adaptive Gaussian attention mask, and (4) provides the attended image to an MLLM for answer generation. \textbf{Our method combines targeted patch selection with dynamic attention to mimic human-like focus on relevant document regions.}}
    \label{fig:method-overview}
\end{figure*}

\noindent At its core, the goal of \methodname is to make MLLMs focus on specific document regions through a query-guided attention mechanism. Given a document image and a query seeking fine-grained information, our method first divides the image into a grid of patches and identifies the most relevant patch using semantic similarity between the query and visual content. It then generates an adaptive Gaussian attention mask centered on this region, effectively highlighting the "needle" while maintaining visibility of the surrounding context. This attended image, along with the original query, is then processed by an MLLM to generate the final answer. Figure \ref{fig:method-overview} illustrates this process.

\subsection{Problem Formulation}

The task of finding fine-grained details in documents can be formalized in both closed-domain and open-domain settings. In the closed-domain setting, given a query $q$ and a document $D$ containing a set of page images $\{I_1, ..., I_j\}$, the goal is to locate the specific region within these images that contains the answer to $q$. The open-domain setting extends this to a collection of documents $S = \{D_1, ..., D_M\}$, where we must first identify the relevant documents and pages before locating the specific region. In the open-domain setting, the top-$r$ relevant documents are retrieved using methods such as ColPali~\cite{faysse2024colpali}, and then passed to the MLLM $L$.




Formally, our objective is to learn a function $f$ that maps a query $q$ and a set of $k$ images $\{I_1, I_2, \dots, I_k\}$ to corresponding attention masks $\{M_1, M_2, \dots, M_k\}$ that highlight the regions most likely to contain the answer:

\vspace{-1em}
\begin{equation}
    \{M_1, M_2, \dots, M_k\} = f(q, \{I_1, I_2, \dots, I_k\})
\end{equation}
\vspace{-0.5em}  

The attended images $\{I_{M_1}, I_{M_2}, \dots, I_{M_k}\}$ are then provided to an MLLM $L$ along with the query to generate the answer:

\vspace{-1em}
\begin{equation}
    \text{answer} = L\left(q, \{I_{M_1}, I_{M_2}, \dots, I_{M_k}\}\right)
\end{equation}

The key challenge lies in designing $f$ to effectively identify and highlight small regions containing critical information while maintaining sufficient context for the MLLM to reason about the answer.

\subsection{Method Overview}

\methodname addresses the challenge of fine-grained detail extraction through a modular pipeline that mimics human visual search behavior. As illustrated in Figure \ref{fig:method-overview}, our method consists of two key components:

\noindent \textbf{Query-Guided Patch Identification:} First, we divide the input document image into an $n \times n$ grid of patches. Using a vision-language model (SigLip~\cite{zhai2023sigmoidlosslanguageimage}), we compute semantic similarity between the query and each patch to identify the region most likely to contain the answer. This step is analogous to how humans quickly scan a document to locate relevant sections based on visual and semantic cues.

\noindent \textbf{Adaptive Gaussian Attention:} Once the most relevant patch is identified, we generate a Gaussian attention mask centered on this region. The spread of this Gaussian distribution adapts dynamically based on the confidence of our patch selection - higher confidence leads to more focused attention, while lower confidence results in broader attention. 
This mechanism directs the MLLM's focus to the identified region while preserving awareness of the surrounding context, similar to human attention.

\noindent The final attended image, created by applying this adaptive Gaussian mask to the original document, serves as input to an MLLM along with the original query. This approach enables the model to efficiently process fine-grained details within the highlighted region while maintaining awareness of the document's overall context, leading to more accurate answers for queries about specific details.

\subsection{Query-Guided Patch Identification}
The first key challenge in locating fine-grained information is identifying which region of the document to focus on. Our patch identification approach combines grid-based image segmentation with semantic similarity matching to efficiently locate regions relevant to the query.

\noindent \textbf{Image Segmentation:} Given an input document image $I$ of dimensions $W \times H$, we divide it into an $n \times n$ grid of uniform patches. Each patch $P_{ij}$ ($i,j \in \{1,...,n\}$) represents a distinct region of the document. Through empirical analysis on our benchmark dataset, we found that $n=6$ provides an effective balance between granularity and computational efficiency.

\noindent \textbf{Query-Patch Similarity:} To identify the most relevant patch, we leverage the SigLip vision-language model to compute semantic similarity between the query and each patch. First, we preprocess the query $q$ by removing stop words and extraneous information to obtain a cleaned query $q_c$, focusing on key semantic elements. The SigLip model then encodes both the cleaned query and each patch into embedding vectors:
\setlength{\abovedisplayskip}{0pt} \setlength{\abovedisplayshortskip}{0pt}
\setlength{\belowdisplayskip}{0pt} \setlength{\belowdisplayshortskip}{0pt}

\vspace{-1em}
\begin{equation}
    v_q = \text{SigLip}(q_c), \quad v_{ij} = \text{SigLip}(P_{ij})
\end{equation}

The relevance of each patch to the query is determined by computing the cosine similarity between their respective embeddings:

\vspace{-1em}
\begin{equation}
    \text{Sim}(v_{ij}, v_q) = \frac{v_{ij} \cdot v_q}{\|v_{ij}\| \|v_q\|}
\end{equation}

\noindent \textbf{Patch Selection:} The patch with the highest similarity score is selected as the center for our attention mechanism:

\vspace{-1em}
\begin{equation}
    (i^*, j^*) = \arg\max_{i,j} \text{Sim}(v_{ij}, v_q)
\end{equation}

\noindent To normalize the similarity score of the selected patch, we apply a softmax function over all similarity scores and define the probability \( p \) of the selected patch as:

\vspace{-0.5em}
\begin{equation}
    p = \frac{\exp(\text{Sim}(v_{i^*j^*}, v_q))}{\sum_{i,j} \exp(\text{Sim}(v_{ij}, v_q))}
\end{equation}

\noindent The center coordinates $(x^*, y^*)$ of this patch in the original image space are computed as:

\begin{equation}
    x^* = \frac{(2j^* - 1)W}{2n}, \quad y^* = \frac{(2i^* - 1)H}{2n}
\end{equation}

\noindent This patch identification process effectively narrows down the region of interest while maintaining computational efficiency. The similarity score of the selected patch also serves as a confidence measure that influences the subsequent attention mechanism, allowing our method to adapt its focus based on the strength of the match between query and content.
\subsection{Adaptive Gaussian Attention}
Once we identify the most relevant patch, the next challenge is to create an attention mechanism that effectively highlights this region while preserving contextual information. We achieve this through an adaptive Gaussian attention mask that automatically adjusts its focus based on the confidence of our patch selection.

\noindent \textbf{Dynamic Gaussian Mask:} We generate a Gaussian attention mask centered at the coordinates $(x^*, y^*)$ identified in the previous step. The spread of this Gaussian distribution is controlled by its standard deviation $\sigma$, which we compute adaptively based on the similarity score $p$ of the selected patch:

\vspace{-1em}
\begin{equation}
    \sigma = \frac{0.8}{1 + \exp(-10(p - 0.2))}
\end{equation}

This sigmoid-based formulation ensures that $\sigma$ varies smoothly with our confidence in the patch selection: high similarity scores result in a broader attention mask (large $\sigma$), reflecting our confidence in finding the answer in that region. In contrast, lower similarity scores yield a smaller mask, as we are less certain about the answer's location, and if $\sigma$ falls below 0.2, we omit drawing a patch altogether. The parameters of this function were determined through empirical analysis on validation set of our benchmark dataset and existing datasets (see ``Dynamic Gaussian Sigma Graph'' in Figure~\ref{fig:method-overview}).

\noindent \textbf{Attention Mask Generation:} The Gaussian attention mask $M(x,y)$ \cite{wu2019gpganrealistichighresolutionimage} for each pixel coordinate $(x,y)$ in the image is computed as:

\begin{equation}
    M(x, y) = \exp\left(-\frac{(x - x^*)^2 + (y - y^*)^2}{2\sigma^2}\right)^{0.5}
\end{equation}
The square root operation in the exponent helps create a more gradual falloff in attention, which we found empirically to work better with MLLMs' visual processing capabilities.

\noindent \textbf{Image Enhancement:} The final attended image $I'$ is created by blending the original image with a highlight color using the attention mask:

\begin{equation}
\begin{aligned}
I'(x,y) = & (1 - \alpha M(x,y))I(x,y) \\
          & + \alpha M(x,y)H(x,y)
\end{aligned}
\end{equation}

\noindent where $\alpha$ is a blending factor (set to 0.5 in our experiments) and $H(x,y)$ represents the highlight color. This approach ensures the highlighted region remains readable and distinct.

\noindent The resulting attended image preserves the document's full content while drawing the MLLM's attention to the region most likely to contain the answer. This balance between focused attention and context preservation is crucial for accurately answering questions about fine-grained details in complex documents.

\section{\methodname: Experimental Setup}
\vspace{-0.2cm}
\subsection{Experimental Datasets}
\textbf{Existing DocVQA Datasets} We evaluate \methodname on two DocVQA datasets: ArxiVQA \cite{li2024multimodalarxivdatasetimproving} and DUDE \cite{van2023document}. 
For evaluation, we use questions, context images, and gold answers from the ArxiVQA training set (since only the training set is available) and the DUDE development set. Hyperparameters are tuned by randomly selecting 50 questions from each dataset. Our test set includes 500 questions from ArxiVQA and 500 from DUDE.

\noindent \textbf{\datasetname} For the evaluation on \datasetname, we select 937 samples distributed across the following domains: Newspapers (174), Menus (180), Lecture Screenshots (70), Website Screenshots (215), Academic Papers (180), and Magazines (118).

\subsection{\methodname Baselines}
Our approach operates in a training-free, zero-shot setting. We evaluate it against two baseline methods: an Optical Character Recognition (OCR)-based pipeline \cite{mishra2019ocr} and the MLLM-DocVQA approach \cite{cho2024m3docrag}. To ensure a comprehensive evaluation, we utilize three closed-source MLLMs—GPT-4o \cite{openai2024gpt4ocard}, GPT-4o-mini \cite{openai2024gpt4ocard}, and Gemini-1.5-flash \cite{geminiteam2024gemini15unlockingmultimodal}—and two open-source MLLMs—Qwen2-VL 7B \cite{wang2024qwen2vlenhancingvisionlanguagemodels} and Llama-3.2-11B-Vision \cite{grattafiori2024llama3herdmodels}. We additionally assess performance under Chain-of-Thought (CoT) prompting~\cite{wei2022chain}. This diverse selection ensures a broad and representative evaluation across both open-source and closed-source models.
\\
\noindent \textbf{OCR-Based Pipeline}
In this pipeline, text is first extracted from a set of images using OCR \cite{mishra2019ocr}, adapted from MMLongBench \cite{mammlongbench}. The extracted text is then input to the LLM, along with the corresponding question, enabling the LLM to generate an answer.
\\
\noindent \textbf{MLLM-Based DocVQA}
This pipeline utilizes MLLMs as the VQA model, where both the question and the corresponding context images are directly input into the model to generate an answer, as adapted from \citet{cho2024m3docrag}.

\subsection{Evaluation Metrics}
We use Exact-Match (EM), F1-Score \cite{rajpurkar2016squad}, and ANLS Score \cite{biten2019scene} as automatic metrics to assess the correctness of the predicted answers. For ArxiVQA, being a multiple-choice question dataset, we use accuracy as the evaluation metric.

For \datasetname, we also conduct human evaluation on 100 samples, with the assistance of three annotators.

\subsection{Implementation Details}
\textbf{Problem Setting:}
We evaluate our method in both open-domain and closed-domain settings. We use DUDE as closed-domain and convert ArxiVQA to open-domain by collating the context of all instances.

\noindent
\textsl{\underline{Open-Domain:}} The top-$k$ most relevant images are retrieved from the corpus to answer queries, using the ArxivQA dataset.
\\
\noindent
\textsl{\underline{Closed-Domain:}} Queries are answered using a predefined set of images that contain the exact query context, evaluated on the DUDE dataset.
\\
\noindent
\textsl{\underline{Distractor Setting:}} Our benchmark, \datasetname, introduces distractor images to assess model resilience against irrelevant information.

\noindent These diverse settings enable a comprehensive evaluation of our proposed method against baseline models.

\noindent\textbf{Context Images and MMLLMs Used:} We use the same set of images across both OCR and MLLM baselines—either for text extraction or as direct inputs to the language model for answering queries. Additionally, we employ same language models for both OCR-based and image-based inputs to ensure consistency and fair comparison.

\noindent\textbf{\methodname Hyperparameters:} For query cleaning, we employ the same Multi-modal Large Language Models (MLLMs) used in the DocVQA task. The image is segmented into a \(6 \times 6\) grid of patches to determine the regions relevant to the query. The standard deviation \(\sigma\) for the 2D Gaussian spread is selected within the range \([0, 0.8]\), as values exceeding 0.8 encompass a substantial portion of the image, thereby negating the intended effect.

\noindent For visualization, patches are highlighted using Blue color, and alpha blending is applied with a blending factor of \(\alpha = 0.5\). Additionally, we impose a threshold of \(\sigma < 0.2\), ensuring that if the final \(\sigma\) falls below this threshold, no patch is drawn. This prevents visualization in cases where the model’s confidence in patch relevance is insufficient. 

\noindent Experiments were performed using two NVIDIA A30 GPUs (24GB each) and MLLMs inference APIs.
\section{Results and Analysis}
This section is divided into two parts:

\noindent (1) \underline{\textsl{\methodname Evaluation:}} We present the results of \methodname using three closed-source models—GPT-4o, GPT-4o-mini, and Gemini-1.5-Flash—and two open-source models—Llama-3.2-VL-11B and Qwen2-7B on ArxiVQA and DUDE datasets. This is followed by an occlusion sensitivity analysis and a detailed error analysis of \methodname. \\
\noindent (2) \underline{\textsl{\datasetname Evaluation:}} We assess the performance of \datasetname on GPT-4o, GPT-4o-mini, Gemini-1.5-Flash, Qwen2-7B, and human evaluators. This is followed by an error analysis of the \datasetname evaluation.

\vspace{-0.2cm}
\subsection{Evaluation on Document Visual QA}
\label{sec: eval_doc}
\vspace{-0.2cm}
\definecolor{highlight}{HTML}{E3F2FD}

\begin{table}[t] 
\scriptsize  
\captionsetup{font=small}
\centering
\begin{tabular}{l c ccc}
\hline     
\textbf{Methods} & \textbf{ArxiVQA} & \multicolumn{3}{c}{\textbf{DUDE}} \\
\cmidrule(lr){2-2} \cmidrule(lr){3-5} 
 & Acc.($\uparrow$) & EM($\uparrow$) & F1($\uparrow$) & ANLS($\uparrow$) \\ 
\hline
\rowcolor{gray!15}
\multicolumn{5}{c}{\fontfamily{lmss}\selectfont{\textsl{Closed-Source LLMs (zero-shot)}}} \\
\hline
GPT-4o & 0.52 & 0.42 & 0.56 & 0.55 \\
\arrayrulecolor{black!5}\hline
GPT-4o-mini & 0.47 & 0.34 & 0.50 & 0.47 \\
\hline
\arrayrulecolor{black!5}\hline
Gem-1.5-Flash & 0.53 & 0.30 & 0.42 & 0.42 \\
\arrayrulecolor{black}\hline
GPT-4o+OCR & 0.41 & 0.34 & 0.47 & 0.47 \\
\arrayrulecolor{black!5}\hline
GPT-4o+CoT & 0.51 & 0.43 & 0.57 & 0.58 \\
\arrayrulecolor{black}\hline
\rowcolor{highlight}
\textbf{GPT-4o+Ours} & \textbf{0.60} & \textbf{0.45} & \textbf{0.60} & \textbf{0.60} \\
\rowcolor{highlight}
\textbf{GPT-4o-mini+Ours} & 0.52 & 0.41 & 0.55 & 0.52 \\
\rowcolor{highlight}
\textbf{Gem-1.5-Flash+Ours} & 0.54 & 0.34 & 0.47 & 0.45 \\
\arrayrulecolor{black}\hline
\rowcolor{gray!15}
\multicolumn{5}{c}{\fontfamily{lmss}\selectfont{\textsl{Open-Source LLMs (zero-shot)}}} \\
\hline
Llama-3.2-VL-11B & 0.41 & 0.13 & 0.23 & 0.18 \\
\arrayrulecolor{black!5}\hline
Qwen2-7B & 0.44 & 0.21 & 0.32 & 0.28 \\
\arrayrulecolor{black}\hline
Llama-3.2+OCR & 0.38 & 0.05 & 0.19 & 0.08 \\
\arrayrulecolor{black!5}\hline
Llama-3.2+CoT & 0.42 & 0.11 & 0.23 & 0.17 \\
\arrayrulecolor{black}\hline
\rowcolor{highlight}
\textbf{Llama-3.2+Ours} & 0.44 & 0.19 & 0.29 & 0.24 \\
\rowcolor{highlight}
\textbf{Qwen2-7B+Ours} & \textbf{0.44} & \textbf{0.27} & \textbf{0.37} & \textbf{0.32} \\
\hline
\end{tabular}
\caption{\textbf{\methodname evaluation} results compared with baselines adapted from M3DocRAG~\cite{cho2024m3docragmultimodalretrievalneed}. \textbf{Our method outperforms all baselines, including CoT \cite{wei2022chain}.}}
\label{tab:comparison}
\end{table}

\begin{table*}[t]
\footnotesize
\centering
\renewcommand{\arraystretch}{1.2}
\setlength{\tabcolsep}{6pt} 

\resizebox{\textwidth}{!}{
\begin{tabularx}{\textwidth}{
    >{\centering\arraybackslash}p{3.3cm}|
    >{\centering\arraybackslash}p{1.3cm} |
    >{\centering\arraybackslash}p{1.5cm} |
    >{\centering\arraybackslash}p{1.4cm} |
    >{\centering\arraybackslash}p{1.4cm} |
    >{\centering\arraybackslash}p{1.6cm} |
    >{\centering\arraybackslash}p{1.2cm} |
    >{\centering\arraybackslash}p{0.8cm} 
}
\toprule
\textbf{Methods} & \textbf{Menus} & \textbf{Academic Papers} & \textbf{Magazines} & \textbf{Newspaper} & \textbf{Website Screenshots} & \textbf{Lectures} & \textbf{All} \\
\midrule
\multicolumn{8}{c}{\textbf{Exact Match (EM)  ($\uparrow$)}} \\
\midrule
\textbf{GPT-4o}             & 0.33 & 0.41 & 0.55 & 0.28 & 0.42 & 0.26 & 0.38 \\
\textbf{GPT-4o-mini}        & 0.25 & 0.23 & 0.47 & 0.24 & 0.34 & 0.24 & 0.29 \\
\textbf{Gemini-1.5-Flash}   & 0.22 & 0.17 & 0.19 & 0.14 & 0.30 & 0.34 & 0.22 \\
\textbf{Qwen2-7B}           & 0.12 & 0.11 & 0.05 & 0.06 & 0.01 & 0.11 & 0.07 \\
\textbf{GPT-4o + Ours}      & \textbf{0.47} & \textbf{0.51} & \textbf{0.64} & \textbf{0.33} & \textbf{0.46} & 0.29 & \textbf{0.46} \\
\textbf{GPT-4o-mini + Ours} & 0.37 & 0.26 & 0.49 & 0.30 & 0.39 & 0.27 & 0.35 \\
\textbf{Gemini-1.5-Flash + Ours} & 0.35 & 0.23 & 0.20 & 0.16 & 0.34 & \textbf{0.41} & 0.27 \\
\textbf{Qwen2-7B + Ours}    & 0.21 & 0.15 & 0.03 & 0.07 & 0.04 & 0.20 & 0.11 \\

\cmidrule(lr){1-8}  
\multicolumn{8}{c}{\textbf{F1 ($\uparrow$)}} \\
\midrule
\textbf{GPT-4o}             & 0.35 & 0.59 & 0.72 & 0.39 & 0.50 & 0.31 & 0.48 \\
\textbf{GPT-4o-mini}        & 0.25 & 0.38 & 0.62 & 0.35 & 0.42 & 0.32 & 0.38 \\
\textbf{Gemini-1.5-Flash}   & 0.22 & 0.29 & 0.25 & 0.20 & 0.36 & 0.40 & 0.28 \\
\textbf{Qwen2-7B}           & 0.16 & 0.19 & 0.07 & 0.11 & 0.01 & 0.12 & 0.10 \\
\textbf{GPT-4o + Ours}      & \textbf{0.50} & \textbf{0.66} & \textbf{0.77} & \textbf{0.44} & \textbf{0.56} & 0.37 & \textbf{0.56} \\
\textbf{GPT-4o-mini + Ours} & 0.38 & 0.41 & 0.64 & 0.37 & 0.49 & 0.36 & 0.44 \\
\textbf{Gemini-1.5-Flash + Ours} & 0.35 & 0.36 & 0.29 & 0.20 & 0.40 & \textbf{0.47} & 0.34 \\
\textbf{Qwen2-7B + Ours}    & 0.27 & 0.24 & 0.06 & 0.10 & 0.04 & 0.20 & 0.15 \\

\cmidrule(lr){1-8}  
\multicolumn{8}{c}{\textbf{ANLS ($\uparrow$)}} \\
\midrule
\textbf{GPT-4o}             & 0.55 & 0.61 & 0.71 & 0.49 & 0.55 & 0.39 & 0.56 \\
\textbf{GPT-4o-mini}        & 0.35 & 0.44 & 0.64 & 0.45 & 0.47 & 0.37 & 0.46 \\
\textbf{Gemini-1.5-Flash}   & 0.29 & 0.40 & 0.35 & 0.32 & 0.47 & 0.42 & 0.37 \\
\textbf{Qwen2-7B}           & 0.19 & 0.29 & 0.18 & 0.25 & 0.08 & 0.16 & 0.19 \\
\textbf{GPT-4o + Ours}      & \textbf{0.63} & \textbf{0.67} & \textbf{0.78} & \textbf{0.52} & \textbf{0.60} & 0.45 & \textbf{0.62} \\
\textbf{GPT-4o-mini + Ours} & 0.49 & 0.46 & 0.67 & 0.46 & 0.51 & 0.40 & 0.50 \\
\textbf{Gemini-1.5-Flash + Ours} & 0.40 & 0.46 & 0.39 & 0.32 & 0.41 & \textbf{0.49} & 0.40 \\
\textbf{Qwen2-7B + Ours}    & 0.26 & 0.32 & 0.17 & 0.23 & 0.11 & 0.23 & 0.22 \\
\bottomrule
\end{tabularx}
}
\caption{\textbf{\datasetname Performance} across different domains including Newspapers, Website Screenshots, and Lectures. While Spot-IT consistently outperforms baseline models, the overall performance remains modest, highlighting the challenging nature of the benchmark and the need for further research and model improvements.}
\label{tab:dataset_eval_grouped_metrics}
\end{table*}

Table \ref{tab:comparison} presents zero-shot results on ArxiVQA and DUDE, comparing our method \methodname to baselines. \methodname consistently outperforms all baselines, including OCR and CoT, highlighting its effectiveness in efficiently finding the “needle” in the set of images. 
We also test our method with the proposed dataset \datasetname, achieving the best performance across all domains in various MLLM models, shown in Table \ref{tab:dataset_eval_grouped_metrics}.

\noindent \textbf{Additional Results} We present further evaluations on additional DocVQA datasets using GPT-4o, alongside extensive ablation studies and patch count analyses, to demonstrate the robustness and generalizability of our Spot-IT framework. Our experiments show that SigLIP consistently outperforms CLIP for patch-query similarity, and that varying the number of patches reveals an optimal trade-off between performance and generalization. Spot-IT achieves consistent gains across multiple benchmarks—DocVQA, InfoVQA, ChartQA, and MMlongbench-doc—for both short and long documents. Detailed results are provided in Appendix~\ref{sec:Additional Results}.



\subsection{Our \datasetname Evaluation}

\textbf{Automatic Evaluation} Table \ref{tab:dataset_eval_grouped_metrics} shows the evaluation of our proposed dataset \datasetname across SoTA MLLMs using EM, F1, and ANLS. These models exhibit low performance both on the overall benchmark and across individual domains, including Restaurant Menus, Newspapers, Website Screenshots, and Lecture Screenshots. This highlights the need to enhance MLLMs and DocVQA methodologies for locating and reasoning about fine-grained details within documents.

\noindent \textbf{Human Evaluation} We evaluate \datasetname using human performance, achieving 63\% EM and 70\% F1, highlighting significant room for improvement compared to MLLMs (Figure \ref{fig:time_chart} in Appendix).

\subsection{Analysis of \methodname}

For our method, we perform:
a) \textsl{Occlusion Sensitivity Analysis} - to understand model behavior,
b) \textsl{Error Analysis} - to interpret failure cases, and
c) \textsl{Accuracy vs. Latency Trade-off Analysis} - comparing our method with baselines.

\subsubsection*{Sensitivity Analysis}
\vspace{-0.2cm}

Figure \ref{fig:Occlusion} shows the occlusion sensitivity analysis of \methodname on the Qwen2-VL model. By systematically occluding image regions, the analysis identifies areas most influential to the model’s predictions. Details of the occlusion methodology are in Appendix \ref{sec:occ_send_anal}.

\noindent \textsl{Findings:} Our method effectively highlights critical image regions that contribute to the model's predictions. This is validated by the occlusion sensitivity analysis, confirming alignment between our method's attributions and the model's decision-making process.

\subsubsection*{Error Analysis}
\vspace{-0.2cm}
We analyze our method on ArxivQA using GPT-4o on 500 samples, of which 200 were incorrect. We randomly selected 50\% of these errors and categorized them as follows: a) \textsl{Dataset Errors} - 19\%, b) \textsl{Retrieval Errors} - 22\%, c) \textsl{Patch Formation} - 25\%, d) \textsl{Patch Selection} - 26\%, and e) \textsl{MMLLM Fault} - 8\%. For details, refer Section \ref{sec:app_error_anla} in the Appendix.

\subsubsection*{Accuracy vs Latency Trade-off}
\vspace{-0.2cm}
The accuracy-latency trade-off plot compares our method with the baseline using GPT-4o on (a) ArxiVQA, (b) DUDE, and (c) \datasetname, showing a 10-20\% accuracy improvement across all datasets with only an additional latency of approximately 4 seconds (see Figure \ref{fig:time_chart} in Appendix).

\subsection{Analysis of \datasetname}

For \datasetname, we conduct: a) \textsl{Error Analysis}, and b) \textsl{Human Evaluation} to compare accuracy and latency with model predictions.

\subsubsection*{\datasetname Error Analysis} 
\vspace{-0.2cm}
We evaluate the performance of \datasetname on GPT-4o by randomly selecting 20 samples from all 6 domains domain and categorized them as follows: a) \textsl{Incomplete Evidence} - 47 cases, b) \textsl{Hallucinated Evidence} - 28 cases, c) \textsl{Perceptual Error} - 24 cases, d) \textsl{Reasoning Error} - 15 cases, e) \textsl{Irrelevant Answer} - 5 cases, and f) \textsl{Knowledge Lacking} - 1 case.
Refer Section \ref{sec:ext_data_err} in Appendix for details.


\subsubsection*{Human vs Model: Accuracy \& Latency }
\vspace{-0.2cm}
We compare human and model performance on accuracy and latency for \datasetname. While humans achieve higher accuracy, they take significantly more time than models, highlighting the need for improved methodologies to efficiently handle our dataset (see Figure \ref{fig:time_chart_nid} in Appendix).

\section{Conclusion}
\vspace{-0.2cm}
In this paper, we formalize the \textsl{Needle in Images} challenge in DocVQA, focusing on evaluating MLLMs' ability to locate and reason about fine-grained details within complex documents. To address this, we introduce \datasetname, a benchmark specifically designed to assess MLLMs' effectiveness in extracting precise information from visually rich layouts. Our experiments reveal that current MLLMs struggle with accurately locating and extracting answers from such intricate structures. To overcome this, we propose \methodname, which intelligently identifies relevant regions within images, achieving substantial improvements over baseline models across multiple datasets. We believe our findings pave the way for more advanced and efficient DocVQA systems capable of fine-grained detail extraction from complex documents.

\section*{Limitations}
The limitations of our work are as follows: 1) Although our method performs well on existing DocVQA datasets, it struggles with long length documents as LLMs have limitations in processing large documents even after identifying the relevant patch. 2) The performance of our method depends on the current capabilities of LLMs, which may improve over time. 3) While achieving high accuracy, our method incurs slightly higher latency due to Gaussian patch construction. 4) We use SigLip for cosine similarity between document patches and the query using a bag-of-words-like approach, which limits contextual understanding of document structure; future work could explore a customized model for better similarity assessment. 5) Our benchmark has fewer complex reasoning questions, which can be expanded in future iterations.


\section*{Acknowledgement}
We thank Pranoy Panda and other members of the AI Lab at Fujitsu Research of India for their valuable feedback on the manuscript. We are also grateful to the anonymous ARR reviewers, the meta-reviewer, and the ACL program chairs for their insightful comments, which helped us improve the paper.

\bibliography{custom}


\appendix
\clearpage
\section{Appendix}
\label{sec:appendix}

In this section, we provide detailed related work and additional results and analysis that we could not include in the main paper due to space constraints. In particular, this appendix contains the following:

\begin{itemize}[nosep]
    \item \hyperref[sec:extended_related_work]{Extended Related Work}
    \item  \hyperref[sec:Additional Results]{Additional results}
        \item \hyperref[sec:add_fig_tab]{Additional Figures and Tables}
    \item  \hyperref[sec:occ_send_anal]{Occlusion Sensitivity Analysis}
    \item  \hyperref[sec:app_error_anla]{Extended \methodname Error Analysis}
    \item  \hyperref[sec:ext_data_err]{Extended \datasetname Error Analysis}
    \item \hyperref[sec:qual_ex]{\methodname Qualitative Examples}

    \item \hyperref[sec:ex_dataset]{Sample Illustrations from \datasetname}
\label{sec:ex_dataset}

    \item \hyperref[sec:prompts used]{All LLM Prompts Used for Evaluation and Dataset Generation}

\end{itemize}

\subsection{Extended Related Work}
\label{sec:extended_related_work}

\begin{table*}[t] 
\scriptsize 
\centering
\begin{tabularx}{\textwidth}{
    >{\centering\arraybackslash}p{3.6cm} |
    >{\centering\arraybackslash}p{1.2cm} 
    >{\centering\arraybackslash}p{1.2cm} 
    >{\centering\arraybackslash}p{1.2cm} 
    >{\centering\arraybackslash}p{1.2cm} 
    >{\centering\arraybackslash}p{1.3cm} 
    >{\centering\arraybackslash}p{3.0cm} }
\toprule
\textbf{Benchmarks} & \textbf{\# Pages/ Document} & \textbf{Unanswerable Questions} & \textbf{Granular Questions} & \textbf{Document Relevance} & \textbf{Answer Source}  & \textbf{Domains}  \\
\midrule

DocVQA \cite{mathew2021docvqa} & 1 & \xmark & \xmark & \xmark & TXT/L/C/TAB/I & Industry Docs \\
ChartQA \cite{masry2022chartqa} & 1  & \xmark & \xmark & \cmark & C & Statista, Pew, OWID, OECD\\
InfoVQA \cite{mathew2022infographicvqa} & 1.2  & \xmark & \xmark  & \xmark & L/C/TAB/I & Infographics Browsing  \\
TAT-DQA \cite{zhu2022towards} & 1.1 & \xmark & \xmark & \xmark & TXT/TAB & Finance Reports \\
DUDE \cite{van2023document} & 5.7  & \cmark & \xmark & \xmark & TXT/L/C/TAB/I &  Books, Media, Public Docs \\
MP-DocVQA \cite{tito2023hierarchical} & 8.3 & \xmark & \xmark & \xmark & TXT/L/C/TAB/I &  Industry Docs \\
ArxiVQA \cite{li2024multimodalarxivdatasetimproving} & 1 & \xmark & \xmark & \xmark & L/C/I & Scientific papers\\
SlideVQA \cite{tanaka2023slidevqa} & 20 & \xmark & \xmark & \xmark & TXT/L/C/TAB/I & SlideDecks  \\
MMLONGBENCH-DOC \cite{mammlongbench} & 47.5  & \cmark & \xmark & \cmark & TXT/L/C/TAB/I & Research and Financial Reports, Academic Papers, Industry Files  \\
\hline
\midrule
\textbf{\datasetname (Ours)} & 29 & \cmark & \cmark & \cmark & TXT/L/C/TAB/I & Menus, Academic Papers, Magazines, Website SS, Lectures SS, Newspapers \\

\bottomrule
\end{tabularx}
\caption{Comparison of benchmarks based on document-level attributes and question types. SS is Screenshots}
\label{tab:dataset_comparison}
\end{table*}

\begin{figure*}[t]
    \centering
    \includegraphics[width=\textwidth]{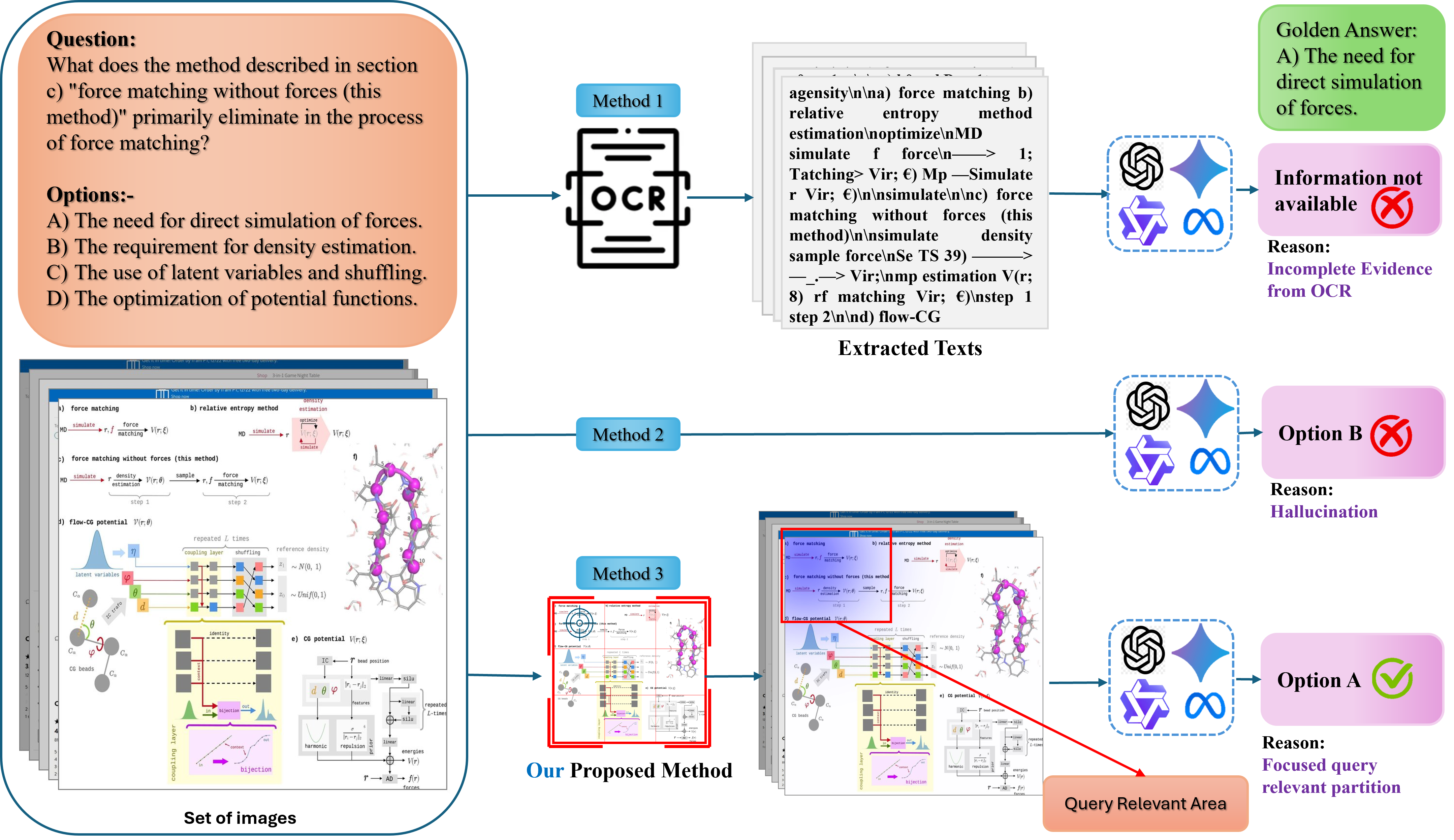}
    \caption{\methodname method comparison with existing methods.\textbf{We highlight failure cases of existing methods and illustrate how \methodname effectively overcomes these challenges.}}
    \label{fig:method-highlevel}
\end{figure*}

\subsubsection{Evolution of Document Visual Question Answering}
Document understanding has evolved significantly from its origins in rule-based systems \cite{srihari1992document} and traditional OCR approaches \cite{subramani2020survey}. Early systems focused primarily on text extraction and basic layout analysis \cite{smith2007overview}, with limited ability to handle complex visual elements or perform sophisticated reasoning. The field has since transformed with the advent of MLLMs \cite{team2023gemini, driess2023palm, peng2023kosmos, openai2023gpt}, which have enabled more nuanced document understanding and reasoning capabilities.

\subsubsection{DocVQA Datasets and Their Evolution}
The development of DocVQA datasets has closely mirrored the advancement in model capabilities. The seminal DocVQA dataset \cite{mathew2021docvqa} established foundational benchmarks for document understanding, focusing primarily on in-line questions where answers could be found within single text spans. This was followed by datasets that introduced additional complexity:

\noindent \textbf{Single-Page Complex Reasoning:} Datasets like CS-DVQA \cite{du2022calm} and RDVQA \cite{wu2022region} pushed beyond simple text extraction by requiring commonsense reasoning and regional understanding. ArxivQA \cite{li2024multimodal} further expanded the challenge by incorporating multiple-choice questions based on academic documents with mixed elements like tables, figures, and charts.

\noindent \textbf{Multi-Page Understanding:} The introduction of multi-page datasets marked a significant evolution in the field. SlideVQA \cite{tanaka2023slidevqa} pioneered questions spanning multiple presentation slides, while MP-DocVQA \cite{tito2023hierarchical} extended document coverage to up to 20 pages. DUDE \cite{van2023document} enriched the challenge by introducing diverse answer types, including lists and arithmetic problems. SPIQA \cite{pramanickspiqa} specifically targeted academic content, requiring sophisticated understanding of scientific figures and plots.

\noindent \textbf{Long-Form Document Understanding:} As MLLMs demonstrated increasing capability in handling standard DocVQA tasks, more challenging benchmarks emerged. MMLongBench-Doc \cite{mammlongbench} represents the current frontier, testing models' ability to reason over long-form documents with complex, multi-step questions. However, none of these datasets specifically target the challenge of locating and reasoning about minute details within larger document contexts—the gap our \datasetname aims to address.

\subsubsection{Methods in Document Understanding}
The methodological approach to document understanding has seen several paradigm shifts:

\noindent \textbf{OCR and Layout-Aware Models:} Early approaches relied heavily on OCR-based pipelines \cite{subramani2020survey}, treating text and visual elements separately. The introduction of layout-aware models like LayoutLM and its variants \cite{xu2020layoutlm, xu2020layoutlmv2, huang2022layoutlmv3} marked a significant advance by incorporating spatial information and document structure into the modeling process.

\noindent \textbf{End-to-End Multimodal Models:} The emergence of powerful MLLMs \cite{team2023gemini, driess2023palm, peng2023kosmos, openai2023gpt} has enabled end-to-end document understanding approaches. Recent methods like CREAM \cite{zhang2024cream} and CFRET \cite{zhang2024cfret} have demonstrated strong performance across various DocVQA tasks.

\noindent \textbf{Retrieval-Augmented Generation:} For larger documents, retrieval-augmented generation (RAG) has emerged as a crucial technique. Methods like ColPali \cite{faysse2024colpaliefficientdocumentretrieval} and M3DocRAG \cite{cho2024m3docrag} have shown promise in efficiently handling large document collections. However, these approaches often process entire document regions without considering information granularity, leading to inefficiencies when answers lie in small, specific regions.

\noindent \textbf{Figure \ref{fig:method-highlevel} shows a comparison of our method, \methodname, with existing methods.}

\subsubsection{Fine-Grained Visual Analysis and Attention Mechanisms}
While fine-grained visual analysis has been extensively studied in natural images, its application to documents presents unique challenges:

\noindent \textbf{Visual Prompting:} Recent work in visual prompting \cite{wu2024visualpromptingmultimodallarge} has shown promising results in directing model attention. Techniques including bounding boxes \cite{lin2024drawandunderstandleveragingvisualprompts}, markers \cite{shtedritski2023doesclipknowred}, and pixel-level annotations \cite{yang2023finegrainedvisualprompting} have proven effective in natural image understanding tasks.

\noindent \textbf{Document-Specific Challenges:} Documents present unique challenges for fine-grained analysis due to their hierarchical structure, complex layouts, and the need to preserve both spatial and semantic relationships. Our \methodname addresses these challenges through a novel question-guided attention mechanism that adapts visual prompting techniques specifically for document understanding tasks.

\subsection{Additional Results}
\label{sec:Additional Results}
Table~\ref{tab:siglip_vs_clip} presents our systematic optimization of the \methodname framework. In our comparison of CLIP and SigLIP for patch-query similarity, SigLIP consistently outperforms CLIP, achieving an accuracy of 0.59 compared to 0.56. Table~\ref{tab:different_n} reports the effect of varying the number of patches ($N$), showing that accuracy increases with $N$ and peaks at $N{=}7$ (0.61), before slightly declining at $N{=}8$ (0.60). We select $N{=}6$ (0.59 accuracy) to ensure better generalizability, striking a balance between strong performance and avoiding potential overfitting at the peak.

Tables~\ref{tab:combined_additional} present comprehensive evaluation results across multiple document understanding benchmarks. It shows consistent improvements with our method over GPT-4o across DocVQA, InfoVQA, and ChartQA, each evaluated on 200 representative questions. For long document understanding, MMlongbench-doc results—evaluated on 54 samples across two runs—further validate the effectiveness of our approach, showing improvements across all metrics: Exact Match, F1, and ANLS.\footnote{We used fewer samples for MMlongbench-doc due to the high computational cost associated with long documents. For similar reasons, we restricted our evaluation to the GPT-4o model across all datasets.}

\begin{table*}[t]
\centering
\resizebox{\textwidth}{!}{%
\begin{tabular}{lcccccccccccc}
\toprule
\multirow{2}{*}{\textbf{Method}} &
\multicolumn{3}{c}{\textbf{DocVQA}} &
\multicolumn{3}{c}{\textbf{InfoVQA}} &
\multicolumn{3}{c}{\textbf{ChartQA}} &
\multicolumn{3}{c}{\textbf{MMlongbench-doc}} \\
& \textbf{EM} & \textbf{F1} & \textbf{ANLS} & \textbf{EM} & \textbf{F1} & \textbf{ANLS} & \textbf{EM} & \textbf{F1} & \textbf{ANLS} & \textbf{EM} & \textbf{F1} & \textbf{ANLS} \\
\midrule
GPT-4o & 0.70 & 0.85 & 0.73 & 0.50 & 0.54 & 0.49 & 0.29 & \textbf{0.32} & \textbf{0.29} & 0.34 & 0.44 & 0.42 \\
GPT-4o + Our Method & \textbf{0.73} & \textbf{0.88} & \textbf{0.74} & \textbf{0.52} & \textbf{0.56} & \textbf{0.50}& \textbf{0.30} & \textbf{0.32} & \textbf{0.29} & \textbf{0.37} & \textbf{0.48} & \textbf{0.45} \\
\bottomrule
\end{tabular}%
}
\caption{Performance (EM / F1 / ANLS) on DocVQA, InfoVQA, ChartQA, and MMlongbench-doc datasets. }
\label{tab:combined_additional}
\end{table*}

\begin{table}
\centering
\begin{tabular}{@{}lc@{}}
\toprule
\textbf{Model} & \textbf{ACC} \\
\midrule
GPT-4o & 0.53 \\
GPT-4o + Spot-IT(CLIP) & 0.56 \\
GPT-4o + Spot-IT(SigLIP) & \textbf{0.59} \\
\bottomrule
\end{tabular}
\caption{Accuracy scores for ArxivQA to compare Siglip and Clip for similarity matching of patch and query}
\label{tab:siglip_vs_clip}
\end{table}

\begin{table}
\centering
\begin{tabular}{@{}lc@{}}
\toprule
\textbf{Method} & \textbf{Acc} \\
\midrule
GPT-4o & 0.53 \\
Spot-IT + GPT-4o(N=3) & 0.60 \\
Spot-IT + GPT-4o(N=4) & 0.58 \\
Spot-IT + GPT-4o(N=5) & 0.59 \\
Spot-IT + GPT-4o(N=6) & 0.59 \\
Spot-IT + GPT-4o(N=7) & \textbf{0.61} \\
Spot-IT + GPT-4o(N=8) & 0.60 \\
\bottomrule
\end{tabular}
\caption{Effect of Number of Patches (N) on Accuracy Score for ArxivQA}
\label{tab:different_n}
\end{table}




\subsection{Additional Figures and Tables}

\begin{figure}[t]
    \centering
    \includegraphics[width=0.4\textwidth]{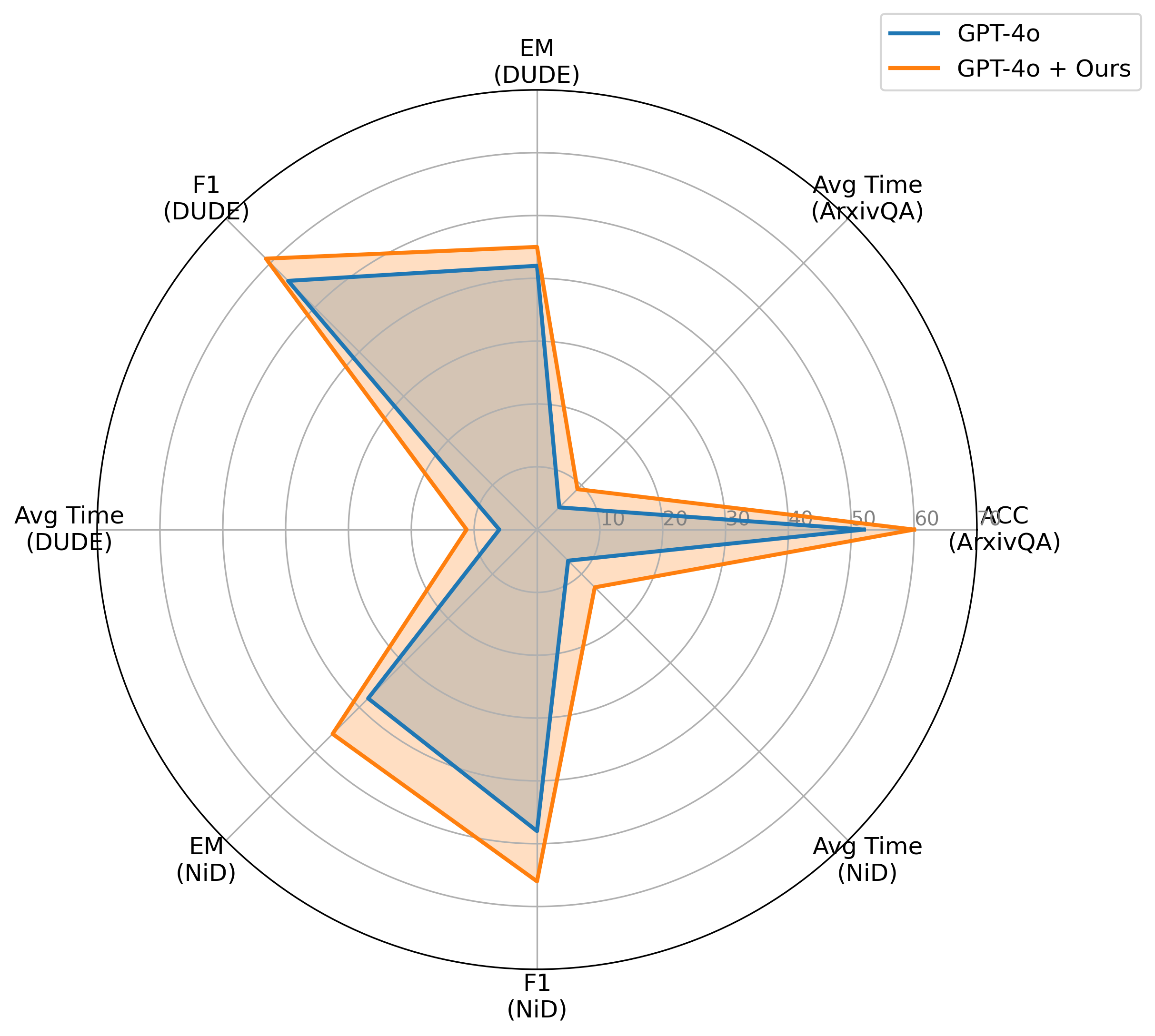}
    \caption{Accuracy and response time comparison of GPT-4o and GPT-4o + Ours on (a) ArxiVQA, (b) DUDE, and (c) \datasetname.}
    \label{fig:time_chart}
\end{figure}

\begin{figure}[t]
    \centering
    \includegraphics[width=0.35\textwidth]{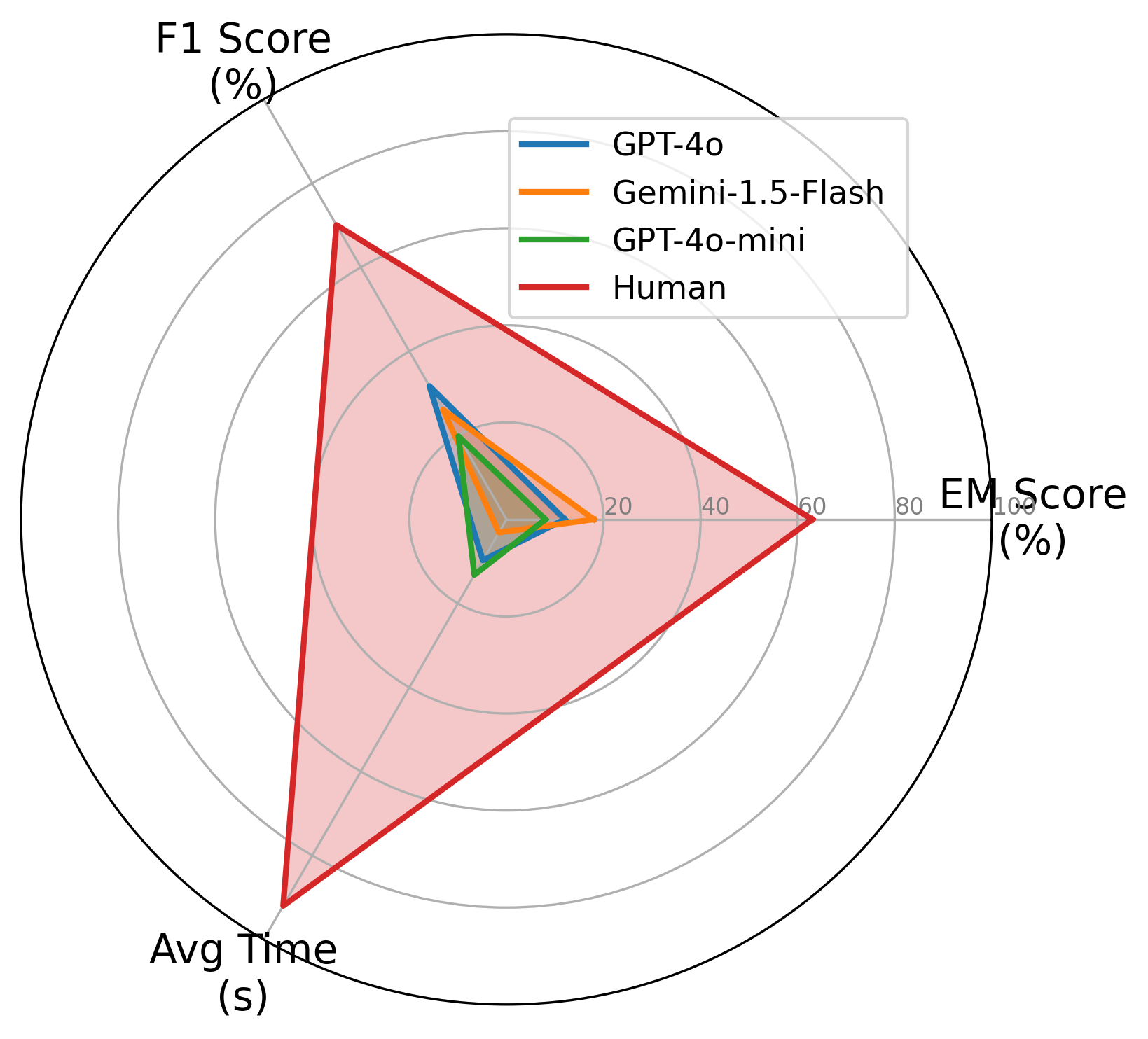}
    \caption{Accuracy and response time comparison on \datasetname (a) for GPT-4o, GPT-4o-mini, Gemini-1.5-Flash, and human.}
    \label{fig:time_chart_nid}
\end{figure}

\begin{enumerate}
    
    \item Table~\ref{tab:dataset-stats} provides a comprehensive overview of the \datasetname dataset. Table~\ref{tab:domain-sources} lists the data sources used to construct \datasetname, while Table~\ref{tab:domain_distribution} outlines the distribution of question categories across various domains. This structured distribution ensures a balanced representation of domain-specific questions, enabling a thorough evaluation of model performance in diverse scenarios.

    \item Table \ref{tab:confusion_matrices} presents results from a Turing test, comparing human-generated and machine-generated responses across different question categories. These results offer insights into the models' capability to generate responses that closely resemble human-like reasoning and linguistic patterns. 

    \item Figure \ref{fig:time_chart} illustrates a comparative performance analysis between GPT-4o and its enhanced variant (GPT-4o + Ours) across multiple well-established benchmarks, including ArxiVQA, DUDE, and NiD-Benchmark. The results demonstrate that \methodname leads to a measurable improvement in accuracy across various tasks. However, this gain comes at the cost of slightly increased inference time, suggesting a trade-off between performance enhancement and computational efficiency.

   \item Figure \ref{fig:time_chart_nid} provides an in-depth examination of the performance gap between AI models and human annotators on the NiD-Benchmark dataset across different domains. The analysis reveals that human responses consistently achieve superior F1 and EM (Exact Match) scores, while also exhibiting a longer average response time. This discrepancy underscores the limitations of existing AI models in achieving human-level comprehension and contextual reasoning, further motivating future advancements in model architectures and training paradigms.
\label{sec:add_fig_tab}
\end{enumerate}

\subsection{Occlusion Sensitivity Analysis}
\label{sec:occ_send_anal}

MLLMs integrate both visual and textual modalities to answer queries about images. Understanding how these models focus on different parts of an image is crucial for interpretability. We implement an occlusion sensitivity method to identify critical image regions that affect model predictions.

\subsubsection{Model and Dataset}

The Qwen2-VL model \cite{wang2024qwen2vlenhancingvisionlanguagemodels} is employed for answering image-based queries. The dataset used is the ArxiVQA dataset..

\subsubsection{Occlusion Sensitivity Analysis}

Given an image $I$ of size $(W, H)$ and a query $Q$, we systematically occlude square patches of the image and measure the change in response probability. The procedure is as follows:

\begin{figure*}[t]
    \centering
    \includegraphics[width=\textwidth]{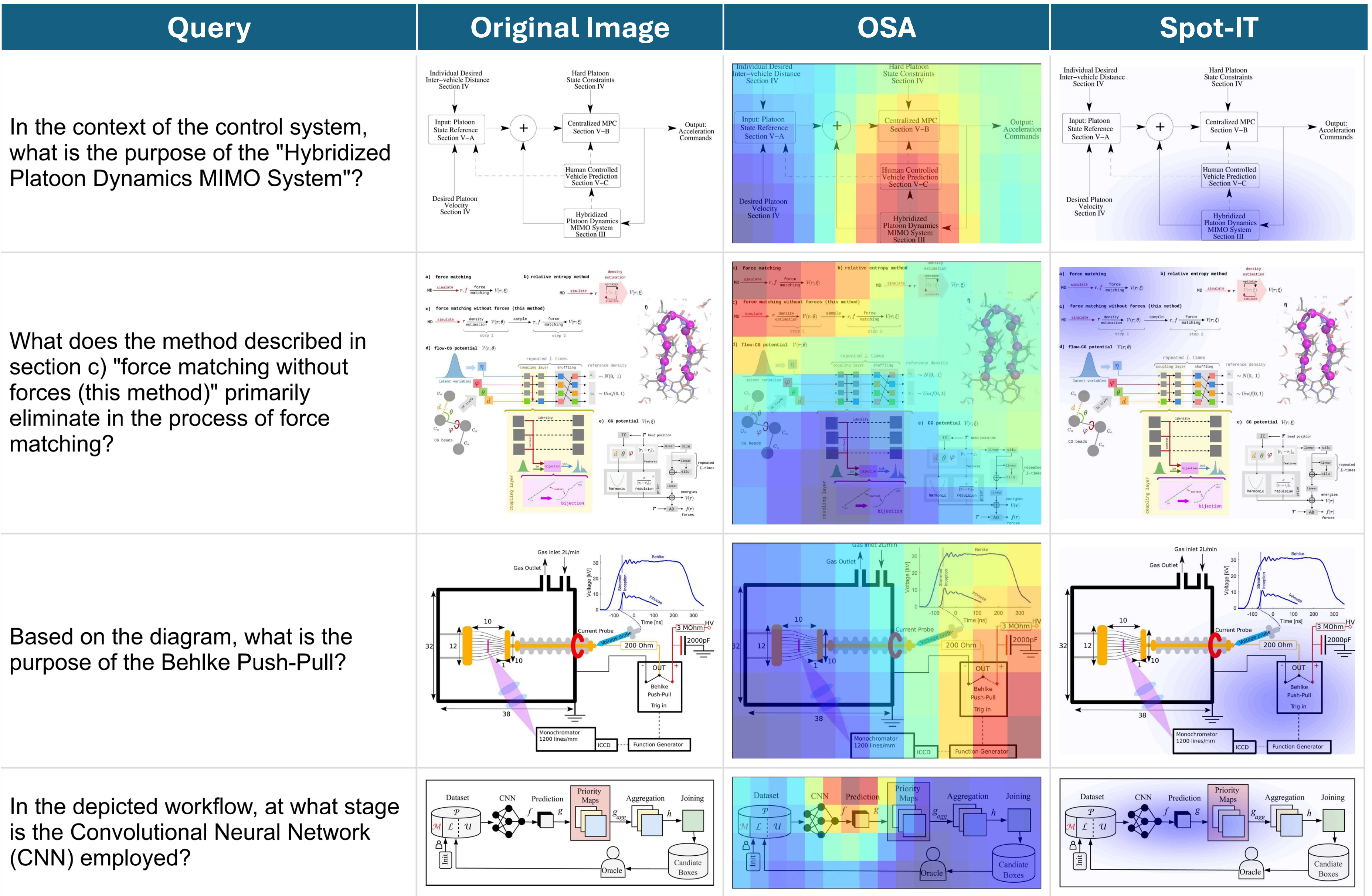}
    \caption{Occlusion Sensitivity Analysis(OSA) comparison with \methodname. \textbf{Demonstrating the correlation between where the MMLLM searches for the answer and where \methodname highlights the images to assist MMLLMs.}}
    \label{fig:Occlusion}
\end{figure*}

\begin{enumerate}[nosep]
    \item Compute the model's original response probability $P_{orig}$.
    \item Slide an occlusion window of size $S \times S$ with stride $T$ over the image.
    \item Replace the windowed region with a neutral color (e.g., black or gray).
    \item Compute the new response probability $P_{occ}$ after occlusion.
    \item Compute the sensitivity score as:
    \begin{equation}
        S(x, y) = P_{orig} - P_{occ}
    \end{equation}
    where $(x, y)$ are the coordinates of the occluded patch.
    \item Generate a heatmap from $S(x, y)$ values and apply Gaussian smoothing.
\end{enumerate}

\subsubsection{Probability Calculation}

To determine the probability of a model's response, the output logits are converted into probabilities using the softmax function:

\begin{equation}
    P(y) = \frac{e^{z_y}}{\sum_{i} e^{z_i}}
\end{equation}

where $z_y$ is the logit corresponding to the generated response.

\subsection{Extended \methodname Error Analysis}
\label{sec:app_error_anla}

We analyze our method on ArxivQA using GPT-4o 
on 500 samples, where 200 samples were incorrect. 
We randomly selected 50\% of these samples and
categorized the errors as follows: 

\begin{itemize}[nosep]

    \item \textbf{Dataset Error (19 cases)}: The dataset had 14 cases of incorrect or ambiguous ground-truth answers, and some questions lacked the necessary context, leading to unavoidable evaluation errors.
    \item \textbf{Retrieval Error (22 cases)}: The retrieval module \cite{faysse2024colpali} failed to fetch relevant information, leading to incorrect answers.

    \item \textbf{Patch Formation (25 cases)}: The patch was incorrectly formed due to a static grid size, leading to improper image cropping and loss of answer context, which caused incorrect matching with the query.

    \item \textbf{Patch Selection (26 cases)}: Incorrect semantic similarity matching occurred between the patch and the input query due to the query's complexity.

    \item \textbf{LLM Fault (8 cases)} Despite having the correct patched image, the Large Language Model sometimes fails to provide the correct answer, particularly for complex questions.
\end{itemize}

\subsection{Extended \datasetname Error Analysis}
\label{sec:ext_data_err}

We evaluate the performance of \datasetname on GPT-4o by randomly selecting 20 samples from all 6 domains domain and categorized them as follows:

\begin{itemize}[nosep]
\setlength\itemsep{0em}
    \item \textbf{Incomplete Evidence (47 cases)}: MLLM is not able to find an evidence to answer the question.
    
    \item  \textbf{Hallucinated Evidence (28 cases)}:MLLM is either answering unanswerable questions or hallucinating the response.
    
    \item \textbf{Perceptual Error (24 cases)}: MLLMs struggle to perceive details such as incorrect decimal placements, leading to inaccurate answers.
    
    \item \textbf{Reasoning Error (25 cases)}: MLLMs struggle to reason accurately, often selecting the first piece of evidence in the relevant section without verifying its correctness.
    
    \item \textbf{Irrelevant Answer (5 cases)}: MLLM is not able to reason deeply and relies on pattern matching, leading to irrelevant answers. It often prioritizes the most prominent or recent context, resulting in inaccurate responses.
    
    \item \textbf{Knowledge Lacking (1 case)}: MLLMs may lack knowledge due to outdated training data, insufficient domain-specific information, or limited context understanding. Additionally, they may struggle with complex reasoning or nuanced details not well-represented in the training corpus.

\end{itemize}


\begin{table}[htbp]
\centering
\small 
\begin{tabular}{lrlr}
\toprule
\multicolumn{4}{c}{\textbf{Statistics}} \\ 
\midrule
Domains & 6 & Categories & 6 \\
Newspapers & 22 & Academic Papers & 32 \\
Magazines & 17 & Lecture Shots & 50 \\
Web Shots & 100 & Menus & 60 \\
Pages/Images & 2,970 & Questions & 1,180 \\
\midrule
\multicolumn{2}{c}{\textbf{Question Statistics}} & \multicolumn{2}{c}{\textbf{Answer Statistics}} \\
\cmidrule(r){1-2} \cmidrule(l){3-4}
Max Length & 26 & Max Length & 19 \\
Avg Length & 10.96 & Avg Length & 1.92 \\
\bottomrule
\end{tabular}
\caption{Dataset Statistics for \datasetname}
\label{tab:dataset-stats}
\end{table}

\begin{table*}[htbp]
\centering
\small 
\begin{tabular}{l l}
\toprule
\textbf{Domain} & \textbf{Source} \\
\midrule
Restaurant Menus &
Various Sources including
\href{https://www.heathrow.com/at-the-airport/restaurants-a-z?type=restaurant}{Heathrow Restaurants},  
\href{https://www.stanstedairport.com/at-the-airport/restaurants/}{London Stansted  Restaurants} etc.  
\\
Academic Papers  & \href{https://arxiv.org/}{Arxiv (2024-2025)} \\
Magazines & \href{https://freemagazines.top/}{freemagazines.top}  \\
Newspapers & \href{https://www.dailyepaper.net/the-times-of-india-epaper/\#download-now-links-of-The-Times-of-India-ePaper}{Times of India},  
\href{https://epaper.thehindu.com/Home/DirectSubscription?tpcc=THEPBG\&msclkid=b4eca5e7c67c12961fc0c245137243bc\&utm_source=bing&utm_medium=cpc&utm_campaign=THEP-Brand-Search\&utm_term=the\%20hindu\%20e-paper\&utm_content=The\%20Hindu\%20ePaper\%20-\%20Exact\%20Match}{The Hindu},  
\href{https://epaper.hindustantimes.com/}{Hindustan Times} (2024-2025) \\
Website Screenshots & CoVA dataset \cite{kumar-etal-2022-cova} \\
Lecture Screenshots & \href{http://ocw.mit.edu/6-034F10}{MIT 6.034 AI, Fall 2010 (MIT OCW)} \\
\bottomrule
\end{tabular}
\caption{Data Sources used to construct the \datasetname dataset across different domains}
\label{tab:domain-sources}
\end{table*}

\begin{table*}[h]
    \centering
    \small
    \renewcommand{\arraystretch}{1.2}
    \begin{tabular}{@{}l c l c l c@{}}
        \toprule
        \textbf{Domain} & \textbf{Count} & \textbf{Domain} & \textbf{Count} & \textbf{Domain} & \textbf{Count} \\
        \midrule
        \textbf{News Paper} &  & \textbf{Lectures} &  & \textbf{Screenshots} &  \\
        \quad Inline        & 199 & \quad Inline        & 48  & \quad Inline        & 203 \\
        \quad Comparative   & 10  & \quad Comparative   & --  & \quad Comparative   & -- \\
        \quad Unanswerable  & 7   & \quad Unanswerable  & 15  & \quad Unanswerable  & 3 \\
        \quad Reasoning     & --  & \quad Reasoning     & 25  & \quad Reasoning     & 35 \\
        \quad Boolean       & --  & \quad Boolean       & 12  & \quad Boolean       & 5 \\
        \quad Commonsense   & --  & \quad Commonsense   & 2   & \quad Commonsense   & -- \\
        \quad \textbf{Total} & \textbf{216} & \quad \textbf{Total} & \textbf{102} & \quad \textbf{Total} & \textbf{246} \\
        \midrule
        \textbf{Academic Paper} &  & \textbf{Magazines} &  & \textbf{Menus} &  \\
        \quad Inline        & 185 & \quad Inline        & 180 & \quad Inline        & 143 \\
        \quad Comparative   & 22  & \quad Comparative   & 9   & \quad Comparative   & 21 \\
        \quad Unanswerable  & 8   & \quad Unanswerable  & 3   & \quad Unanswerable  & -- \\
        \quad Reasoning     & 5   & \quad Reasoning     & 10  & \quad Reasoning     & -- \\
        \quad Boolean       & --  & \quad Boolean       & --  & \quad Boolean       & 23 \\
        \quad Commonsense   & --  & \quad Commonsense   & --  & \quad Commonsense   & 7 \\
        \quad \textbf{Total} & \textbf{220} & \quad \textbf{Total} & \textbf{202} & \quad \textbf{Total} & \textbf{194} \\
        \bottomrule
    \end{tabular}
    \caption{\datasetname Distribution of Question Categories Across Domains}
    \label{tab:domain_distribution}
\end{table*}

\begin{table*}[t]
\footnotesize
\centering
\renewcommand{\arraystretch}{1.2} 
\setlength{\tabcolsep}{4pt} 

\begin{tabularx}{\textwidth}{l|*{2}{>{\centering\arraybackslash}X}|*{2}{>{\centering\arraybackslash}X}}
\hline
\multicolumn{1}{l|}{\textbf{Ground Truth}} & \multicolumn{2}{c|}{\textbf{Gemini 2.0 Flash}}  & \multicolumn{2}{c}{\textbf{Human verifier 1}} \\
\cline{2-5}
& \textbf{Predicted Human} & \textbf{Predicted Machine} & \textbf{Predicted Human} & \textbf{Predicted Machine} \\
\hline
Human   & 146 & 54  & 170 & 30 \\
Machine & 143 & 57  & 160 & 40 \\
\hline
Total   & 289 & 111 & 330 & 70  \\
\hline
\end{tabularx}

\begin{tabularx}{\textwidth}{l|*{2}{>{\centering\arraybackslash}X}|*{2}{>{\centering\arraybackslash}X}}
\hline
\multicolumn{1}{l|}{\textbf{Ground Truth}} & \multicolumn{2}{c|}{\textbf{Claude 3.5 Sonnet}}  & \multicolumn{2}{c}{\textbf{Human verifier 2}} \\
\cline{2-5}
& \textbf{Predicted Human} & \textbf{Predicted Machine} & \textbf{Predicted Human} & \textbf{Predicted Machine} \\
\hline
Human   & 181 & 19  & 171 & 29 \\
Machine & 176 & 24  & 162 & 38 \\
\hline
Total   & 357 & 43 & 333 & 76  \\
\hline
\end{tabularx}

\caption{\textbf{Turing Test and LLM as a Judge Results.} We find that the generated questions in our \datasetname are classified as human-generated with a moderately high agreement score}
\label{tab:confusion_matrices}
\end{table*}

\subsection{Sample Illustrations from \datasetname}

Table \ref{tab:vqa_examples} represents examples from \datasetname encompassing multiple domains and categories to support diverse research applications. The dataset integrates visually rich images from domains such as website screenshots, lecture slides, restaurant menus, magazines, newspapers, and research papers. Each instance is categorized into Boolean, unanswerable, common sense, reasoning, comparative, and inline question-answering tasks.
\label{sec:ex_dataset}

\begin{table*}[!t]
\centering
\renewcommand{\arraystretch}{0.9}  
\setlength{\extrarowheight}{0pt}   
\begin{adjustbox}{max height=12cm}  
\begin{tabularx}{\textwidth}{X|p{2cm}|p{3cm}|p{3cm}|X|X} 
\toprule
\textbf{Domain} & \textbf{Category} & \textbf{Image} & \textbf{Region of Interest} & \textbf{Question} & \textbf{Answer} \\
\midrule
\raggedright Website Screen Shot & 
\raggedright Boolean & 
\raisebox{-\totalheight}{\includegraphics[width=\linewidth]{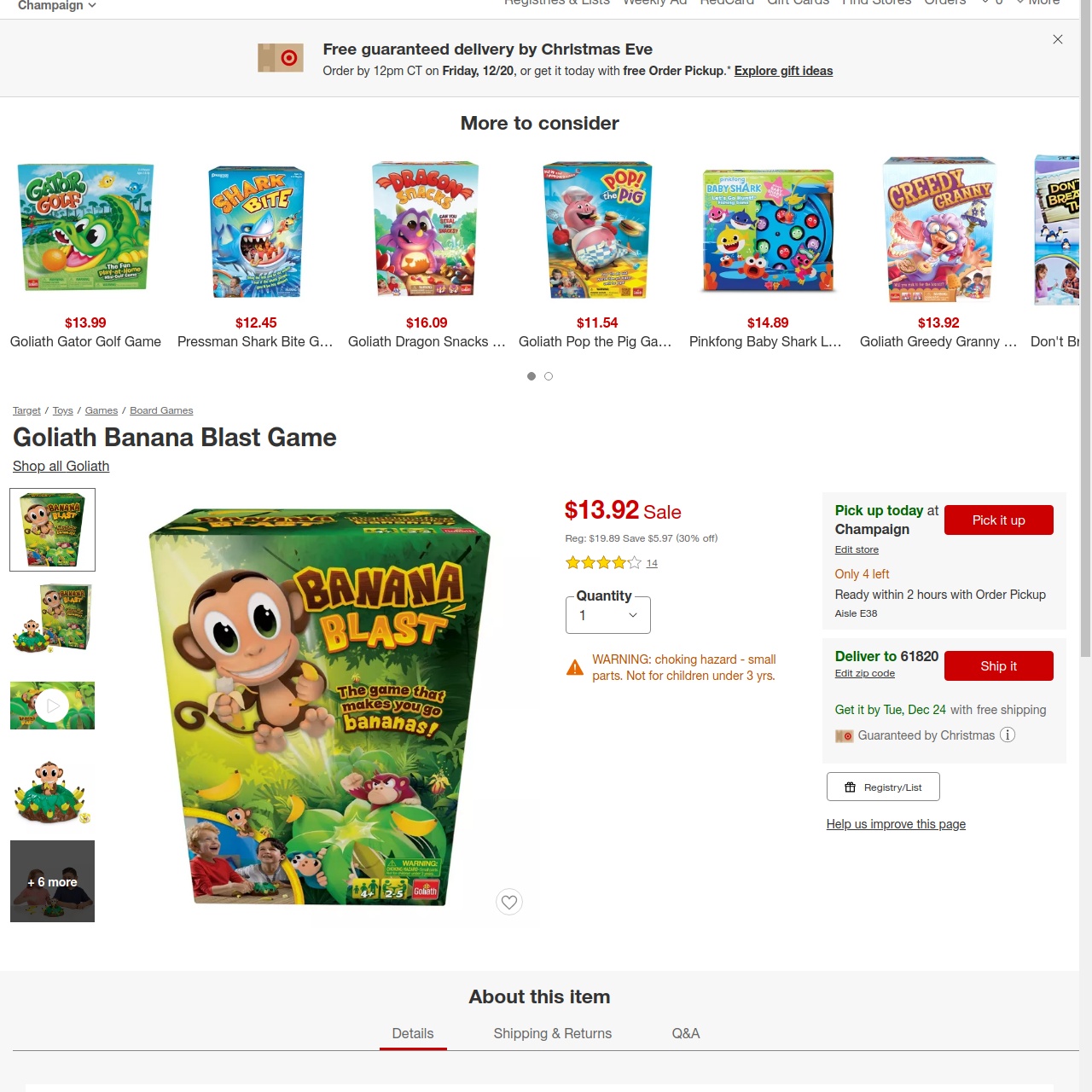}} & 
\raisebox{-\totalheight}{\includegraphics[width=\linewidth]{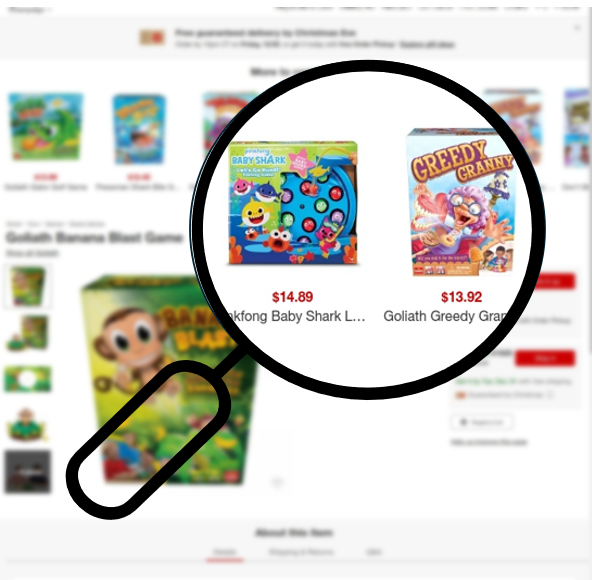}} & 
\raggedright The game "Greedy Granny" and "Baby Shark" are priced the same (True/False)? & 
False \\
\midrule
\raggedright Lecture Screen Shot & 
\raggedright Unanswerable & 
\raisebox{-\totalheight}{\includegraphics[width=\linewidth]{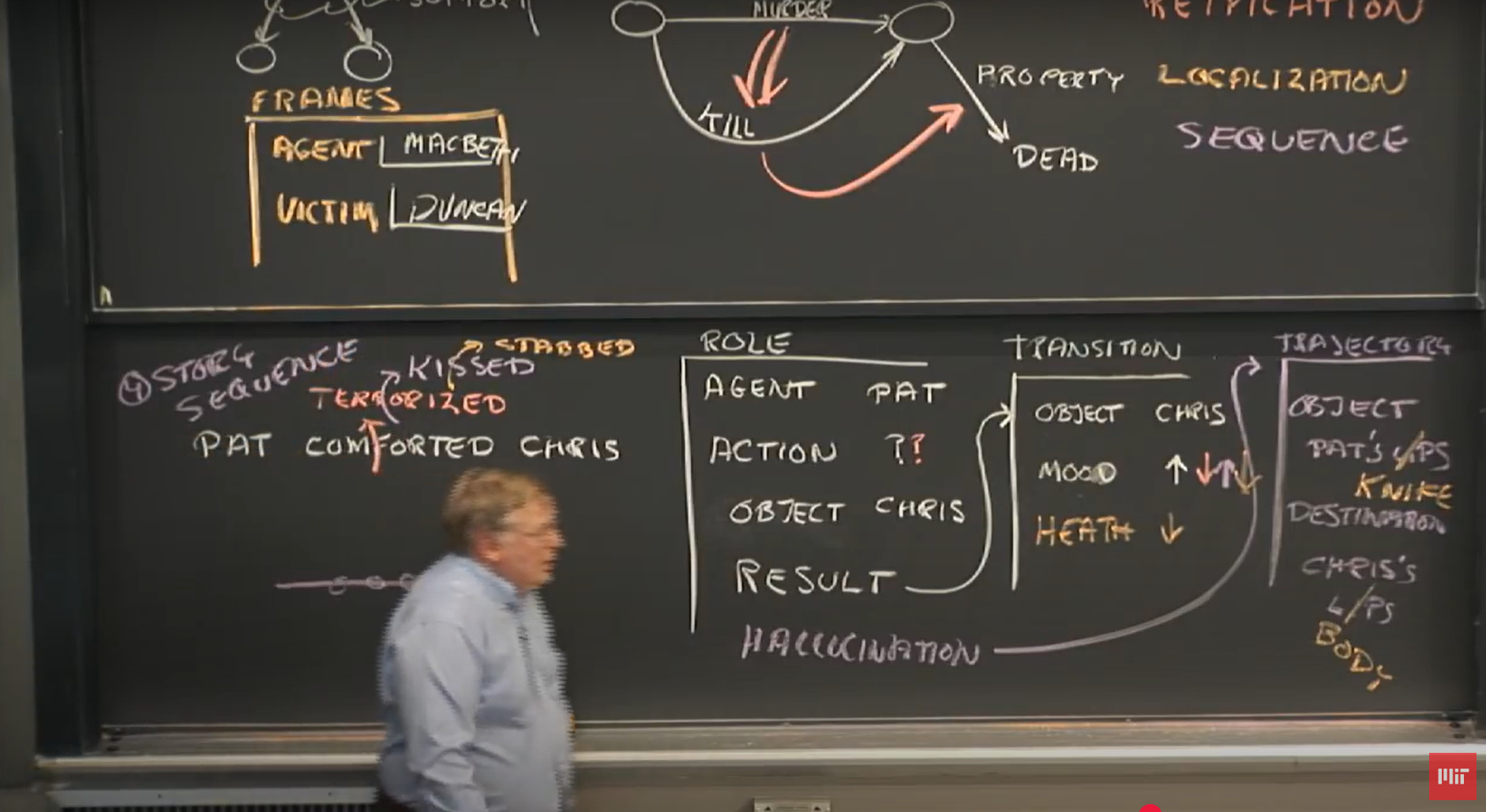}} & 
\raisebox{-\totalheight}{\includegraphics[width=\linewidth]{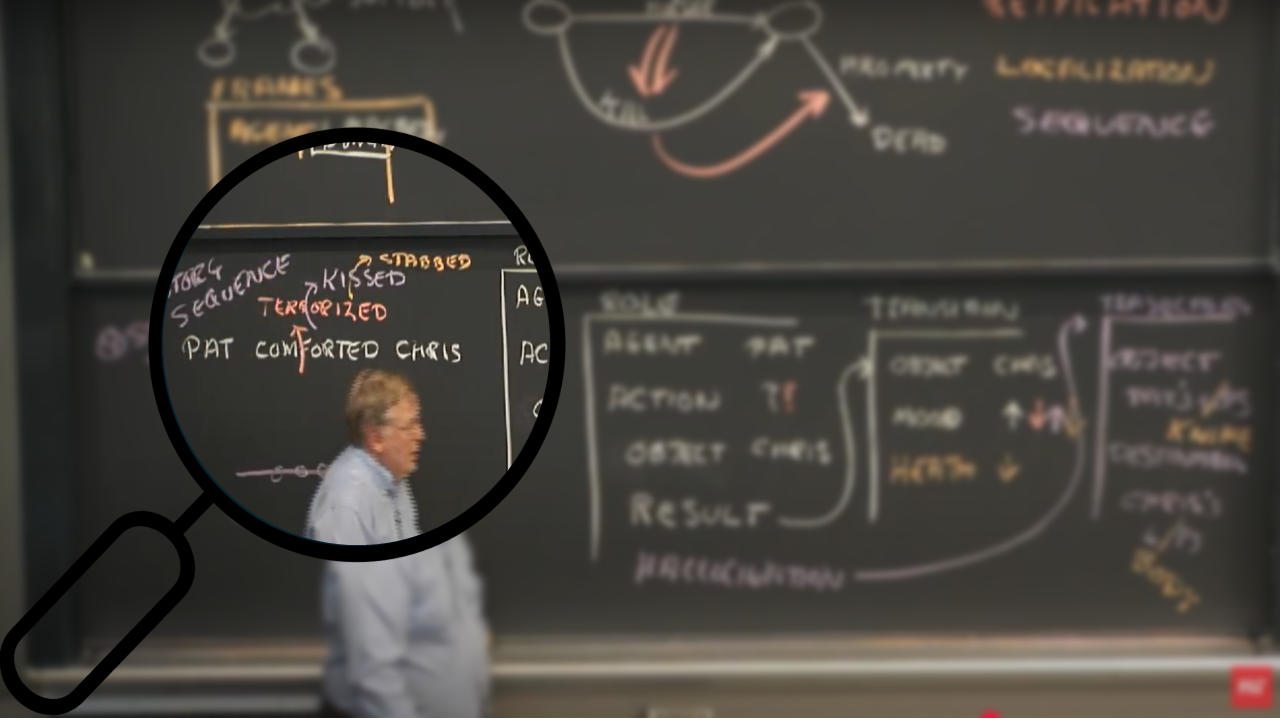}} & 
\raggedright Who Hugged Chris? & 
Information not available \\
\midrule
\raggedright Restaurant Menus& 
\raggedright Common Sense & 
\raisebox{-\totalheight}{\includegraphics[width=\linewidth]{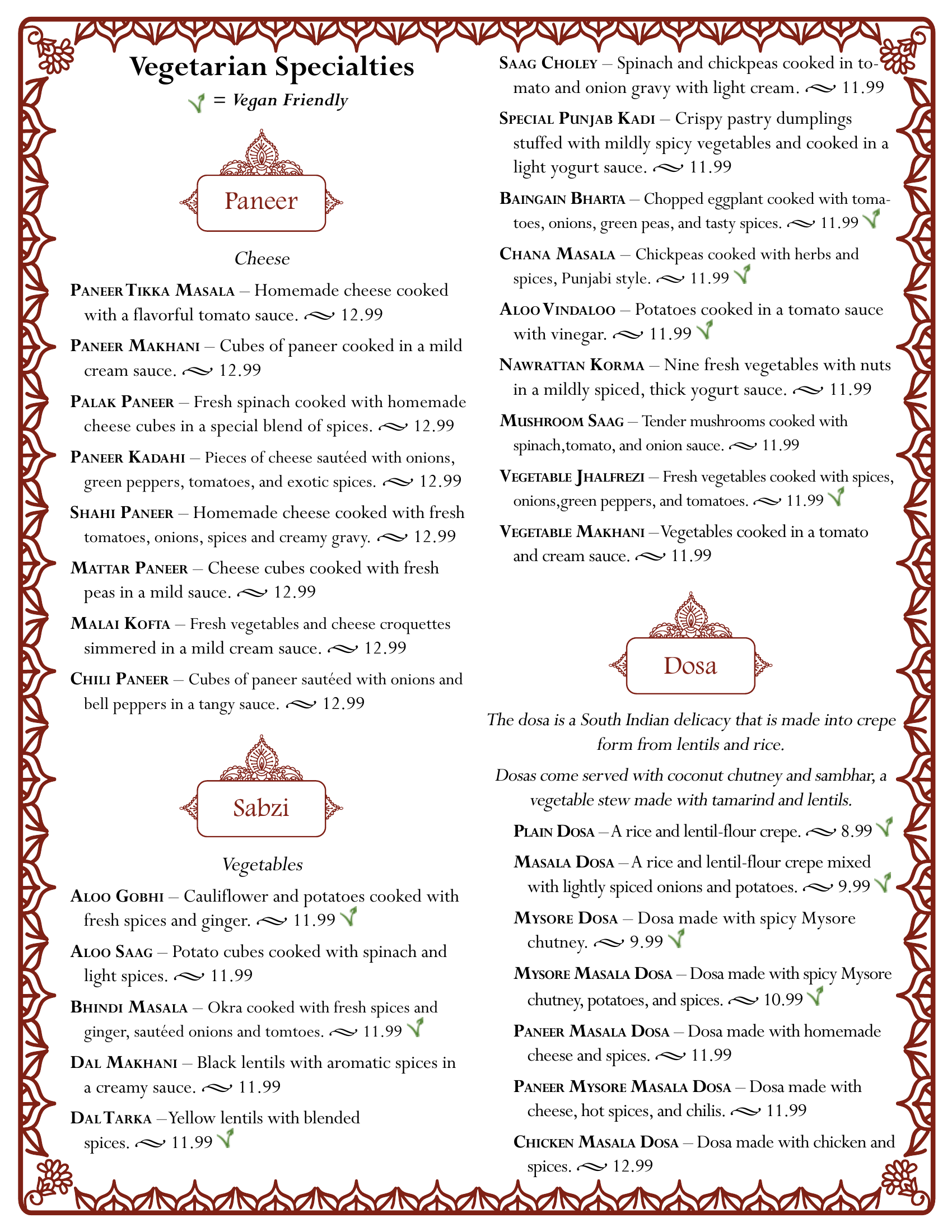}} & 
\raisebox{-\totalheight}{\includegraphics[width=\linewidth]{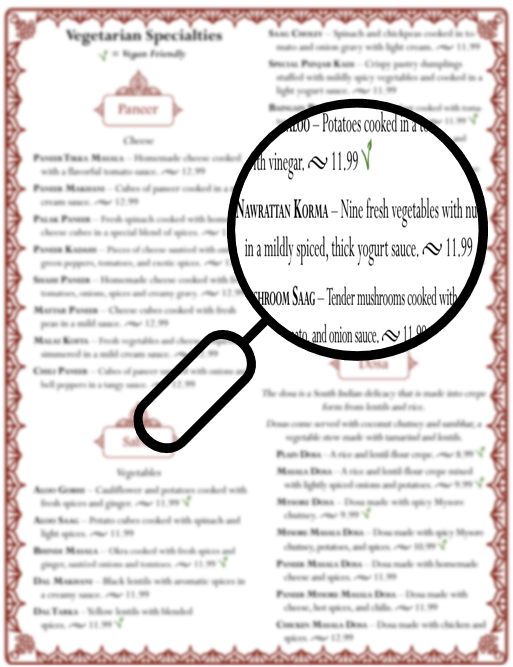}} & 
\raggedright Is the Nawarattan Korma dish vegetarian? & 
Yes \\
\midrule
\raggedright Magazines & 
\raggedright Reasoning & 
\raisebox{-\totalheight}{\includegraphics[width=\linewidth]{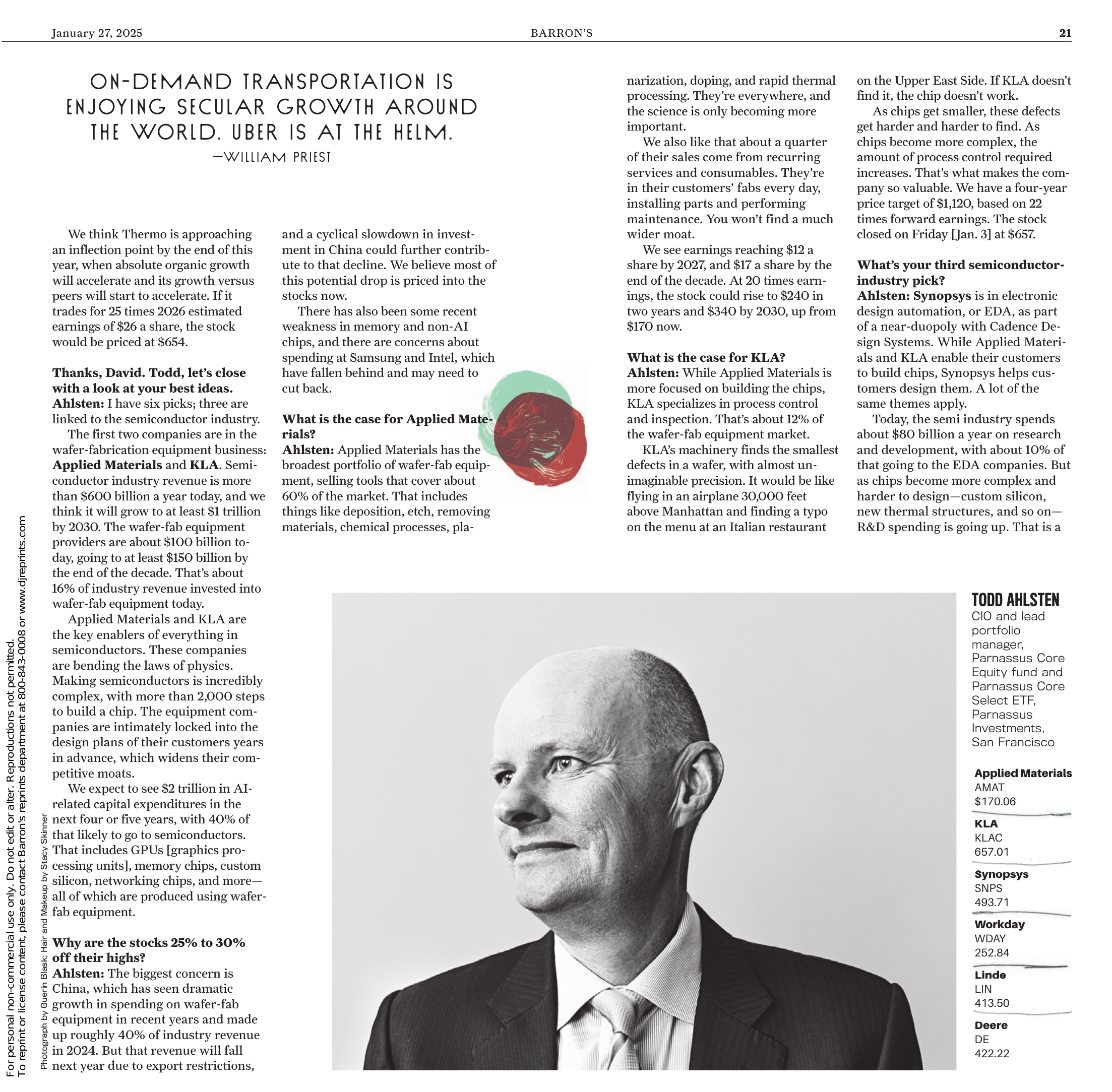}} & 
\raisebox{-\totalheight}{\includegraphics[width=\linewidth]{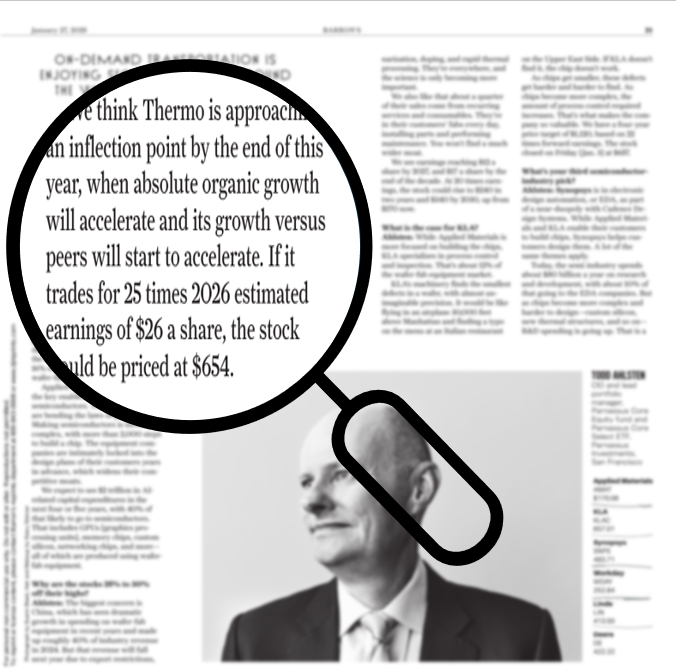}} & 
\raggedright What is the estimated price of Thermo's stock if it trades at 25 times 2026 earnings? & 
\$654 \\
\midrule
\raggedright News Papers & 
\raggedright Comparative & 
\raisebox{-\totalheight}{\includegraphics[width=\linewidth]{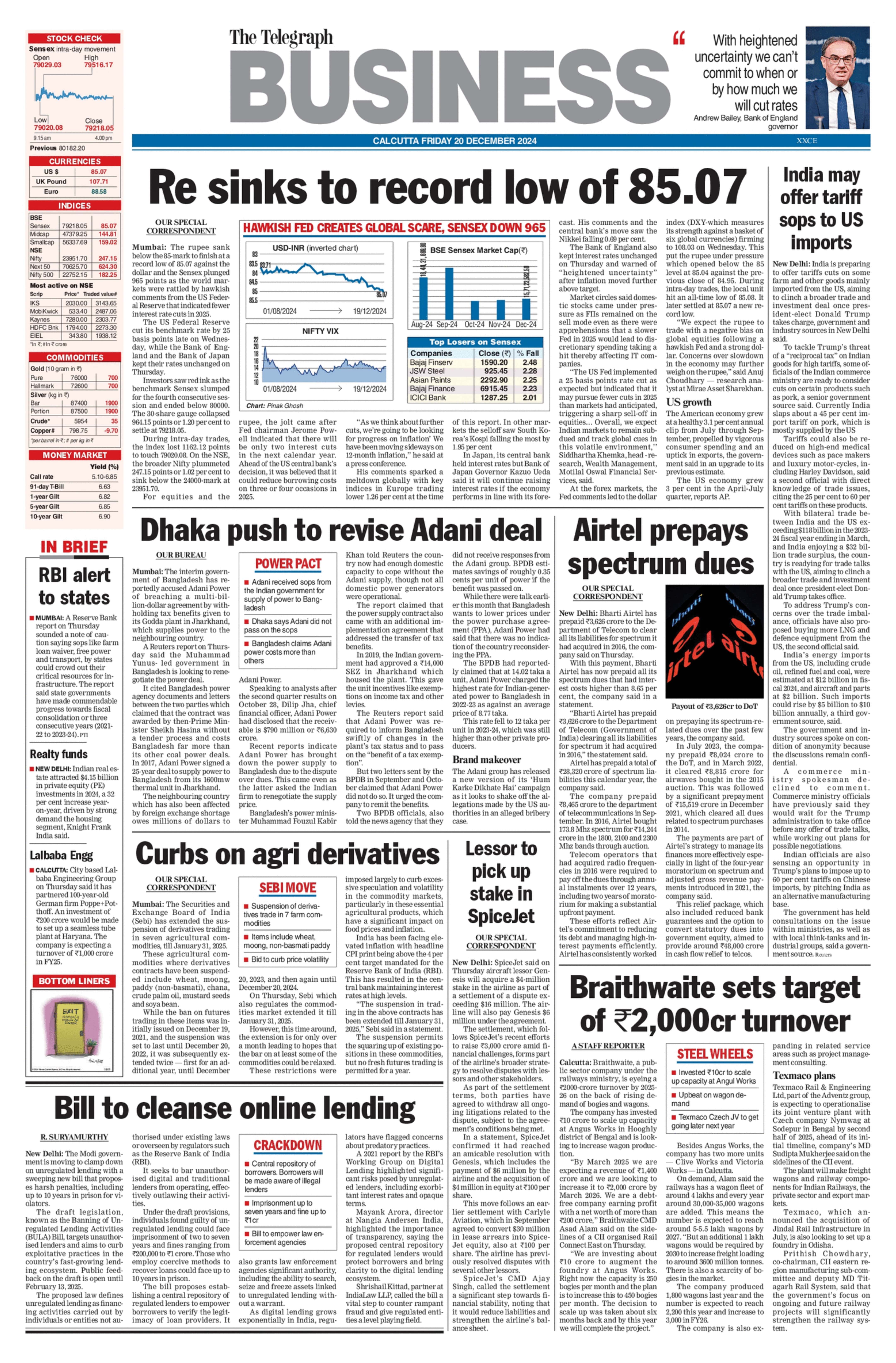}} & 
\raisebox{-\totalheight}{\includegraphics[width=\linewidth]{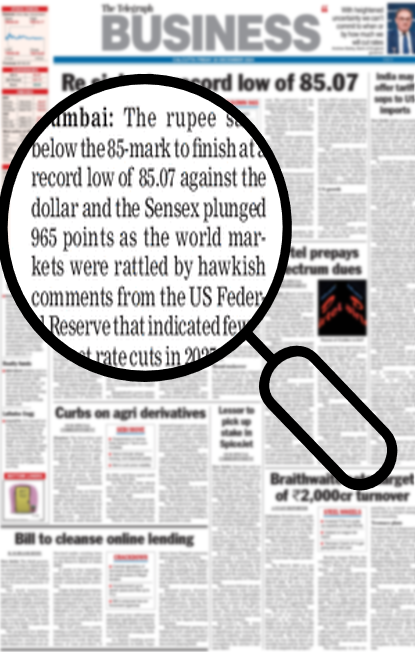}} & 
\raggedright What was the record low value of the rupee against the dollar? & 
85.07 \\
\midrule
\raggedright Research Papers & 
\raggedright Inline & 
\raisebox{-\totalheight}{\includegraphics[width=\linewidth]{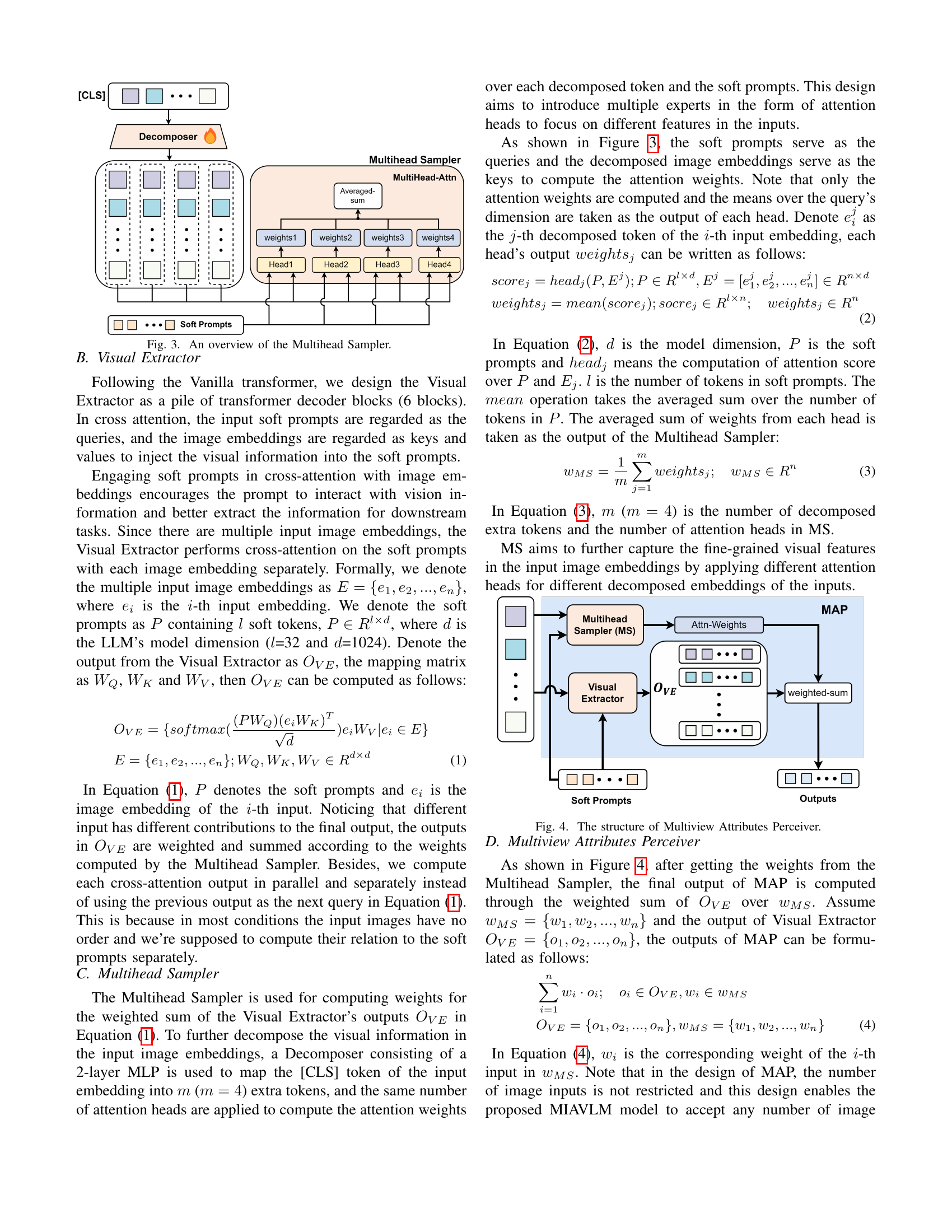}} & 
\raisebox{-\totalheight}{\includegraphics[width=\linewidth]{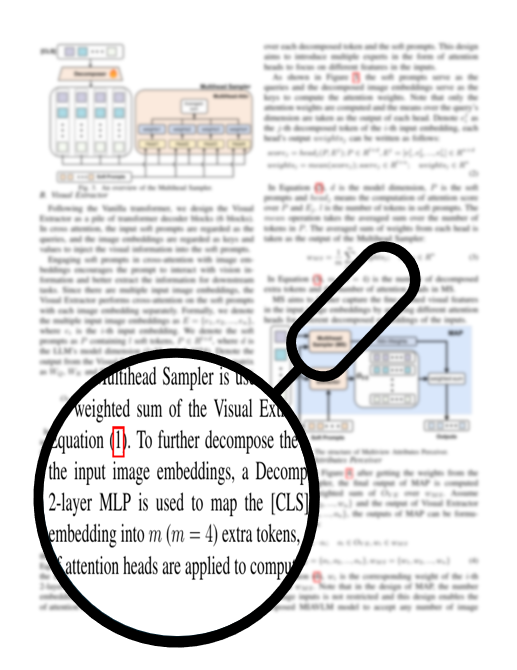}} & 
\raggedright What is the value of m in the Decomposer's MLP? & 
4 \\
\bottomrule
\end{tabularx}
\end{adjustbox}  
\caption{\textbf{Sample Illustrations from \datasetname.} Question-answer pairs across different domains, including the question, required context, question category, and relevant region of interest to find the answer.}
\label{tab:vqa_examples}
\end{table*}

\subsection{\methodname Qualitative Examples}
\label{sec:qual_ex}

\begin{figure*}[htbp]
    \centering
    \includegraphics[width=\linewidth,height = 9cm]{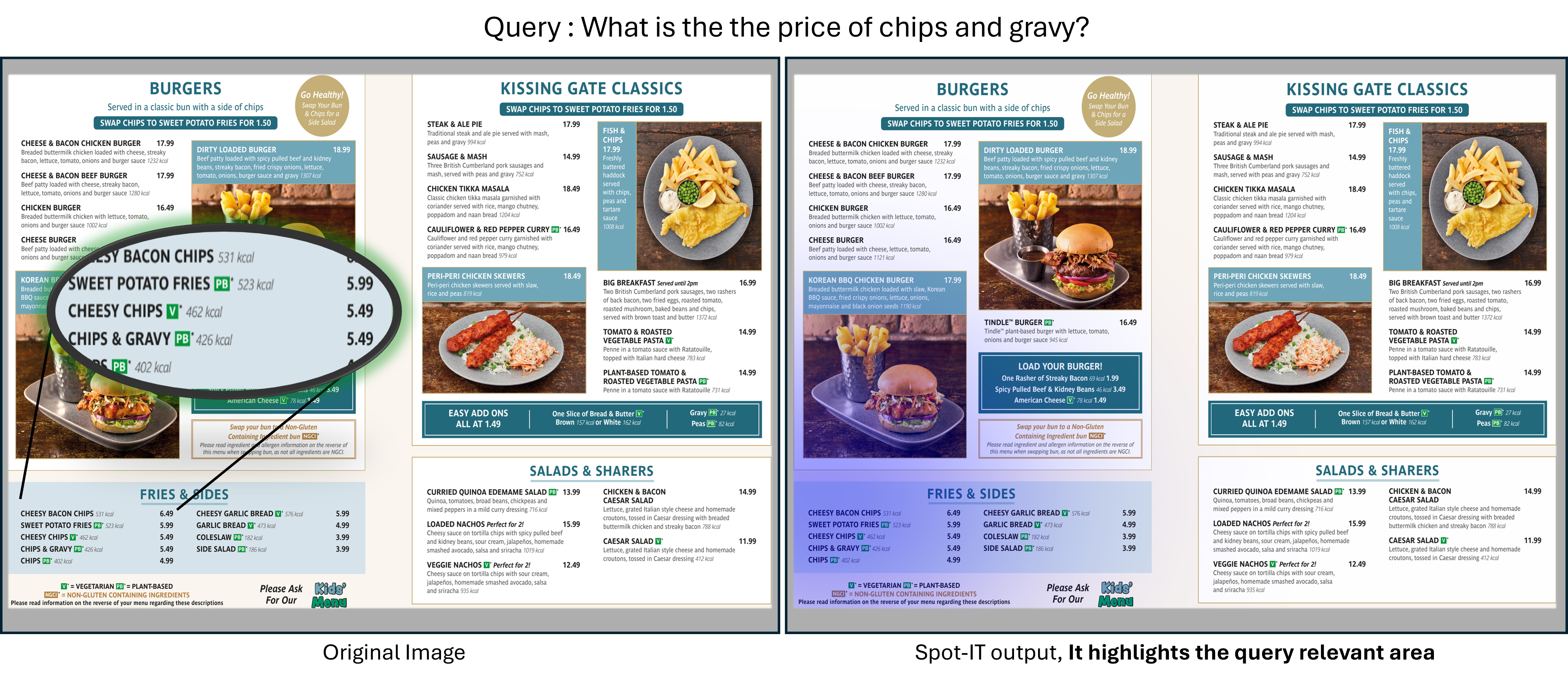}
    \caption{An example from the restaurant menu domain in the NiM benchmark. The Spot-IT method accurately highlights the query-relevant region.}

    \label{fig:qualitative-example-restaurant}
\end{figure*}

\begin{figure*}[htbp]
    \centering
    \includegraphics[width=\linewidth,height = 9cm]{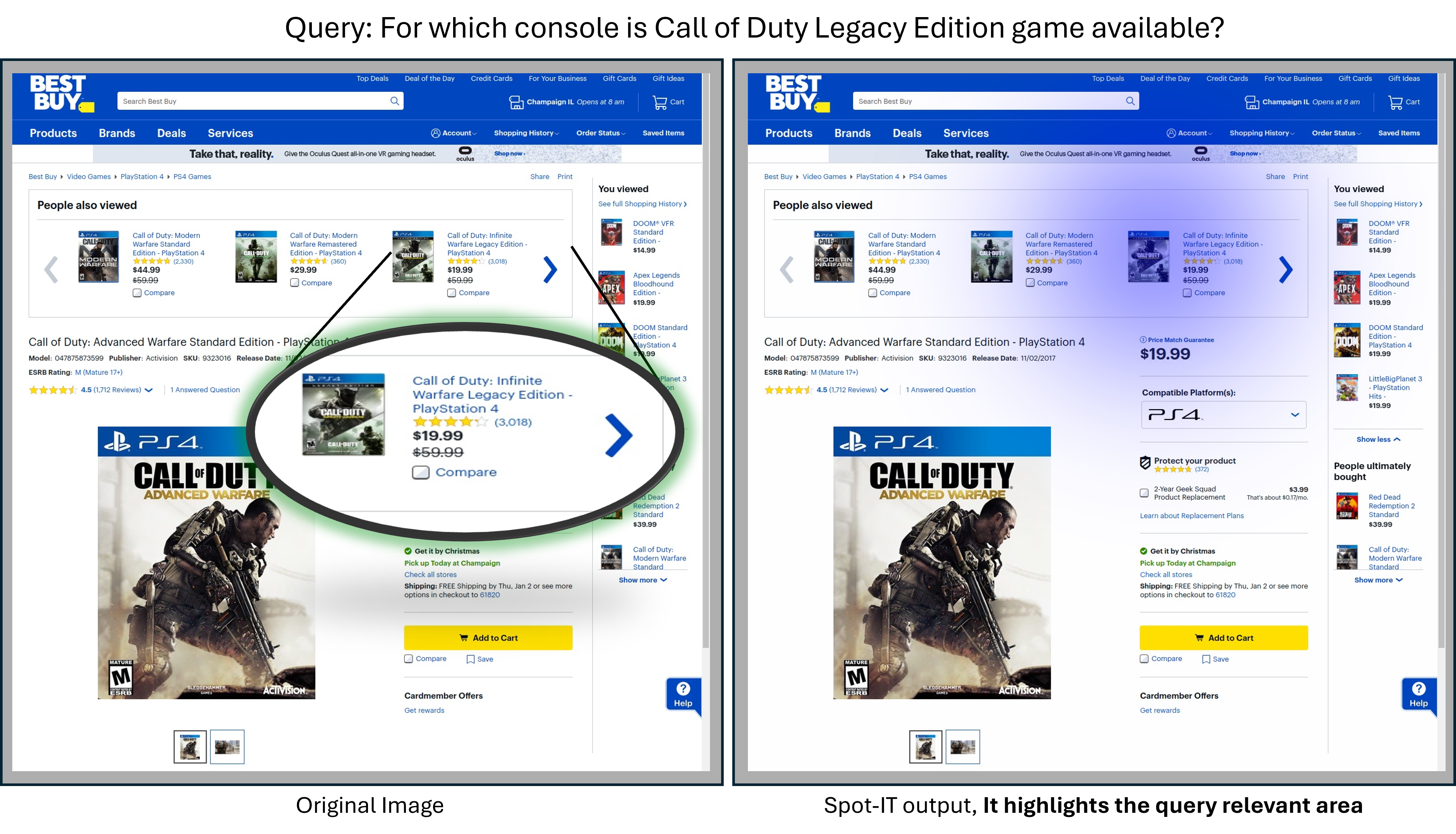}
    \caption{An example from a website screenshot in the NiM benchmark.Spot-IT successfully localizes the visual region relevant to the query.}

    \label{fig:qualitative-example-website ss}
\end{figure*}

\begin{figure*}[htbp]
    \centering
    \includegraphics[width=\linewidth,height = 9cm]{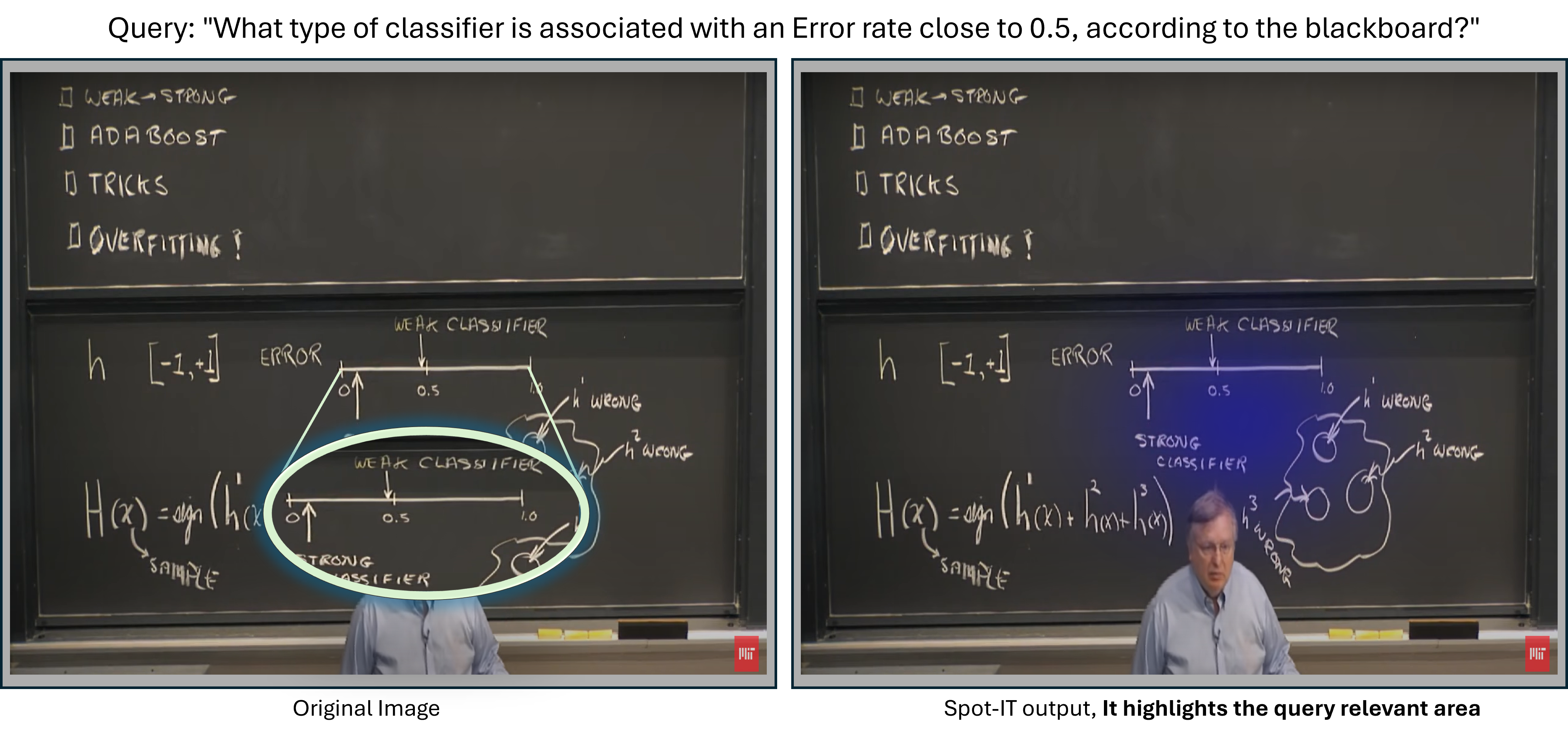}
    \caption{An example from the lecture screen shot domain in the NiM benchmark. Spot-IT effectively highlights the query-specific portion of the blackboard.}

    \label{fig:qualitative-example-lec ss}
\end{figure*}

Figures~\ref{fig:qualitative-example-restaurant},~\ref{fig:qualitative-example-website ss}, and~\ref{fig:qualitative-example-lec ss} present qualitative examples from the NiM benchmark, demonstrating its applicability across diverse domains such as restaurant menus, website screenshots, and lecture slides. These examples emphasize how NiM focuses on fine-grained visual question answering, requiring models to reason over localized and domain-specific visual details. Furthermore, the effectiveness of the proposed Spot-IT method is highlighted, as it successfully identifies and highlights the query-relevant regions in each image. By drawing attention to the most informative parts of the visual input, Spot-IT facilitates better grounding for multimodal large language models, thereby improving their interpretability and VQA performance across different real-world document types.

\onecolumn

\subsection{All LLM Prompts Used for Evaluation and Dataset Generation}
\label{sec:prompts used}

\subsubsection{Prompt for Document VQA Evaluation}
This prompt assesses a model's ability to answer questions based solely on document images, without external knowledge. Responses should be concise (preferably a single word or number). If the information is unavailable, the model should respond with \texttt{"Information not available."}

\subsubsection{Customized Prompt for Document VQA Evaluation}
This variant prioritizes information in blue-highlighted regions, considering the entire image only if necessary. Constraints on external knowledge, concise responses, and handling of missing information remain unchanged.

\subsubsection{QA Generation Prompt for \datasetname}
This prompt generates precise, challenging questions from document images. Each question should be natural, answerable from a small document portion, and uniquely identifiable. Necessary context must be explicit, avoiding vague references.

\noindent Only 2--3 high-quality questions per document should be produced; otherwise, output \texttt{"NA."} The output follows a structured JSON format for consistent benchmarking.


\begin{tcolorbox}[
    enhanced,
    breakable,
    width=\textwidth,
    title=Prompt for Document VQA Evaluation \cite{mammlongbench},
    fonttitle=\bfseries\large,
    colframe=black!75!black,
    colback=white!10!white,
    coltitle=white,  
    colbacktitle=gray,  
    boxrule=0.2mm,
    sharp corners,
    shadow={1mm}{-1mm}{0mm}{black!50!white},
    attach boxed title to top left={yshift=-3mm, xshift=3mm},
    boxed title style={sharp corners, size=small}
]

\textbf{Task:} \\
\textsl{[Images]}\\
\texttt{Read the above Images and answer this question}

\textbf{Instructions:}
\begin{itemize}
    \item \texttt{DO NOT use external knowledge.}
    \item \texttt{Provide a one-word or numerical answer if possible.}
    \item \texttt{If information is unavailable, state "Information not available."}
\end{itemize}

\end{tcolorbox}

\begin{tcolorbox}[
    enhanced,
    breakable,
    width=\textwidth,
    title=Customized Prompt for Document VQA(for \methodname) Evaluation,
    fonttitle=\bfseries\large,
    colframe=black!75!black,
    colback=white!10!white,
    coltitle=white,  
    colbacktitle=gray,  
    boxrule=0.2mm,
    sharp corners,
    shadow={1mm}{-1mm}{0mm}{black!50!white},
    attach boxed title to top left={yshift=-3mm, xshift=3mm},
    boxed title style={sharp corners, size=small}
]

\textbf{Task:} \\
\textsl{[Images]}\\
\texttt{Read the above Images and answer this question}
\\ \\
\texttt{Focus on the BLUE Highlighted area in images as it is more relevant to the query.
                    First, try to answer only using the highlighted area, and if not found, then, consider whole image}

\textbf{Instructions:}
\begin{itemize}
    \item \texttt{DO NOT use external knowledge.}
    \item \texttt{Provide a one-word or numerical answer if possible.}
    \item \texttt{If information is unavailable, state "Information not available."}
\end{itemize}

\end{tcolorbox}

\begin{tcolorbox}[
    enhanced,
    breakable,
    width=\textwidth,
    title=QA Generation Prompt for \datasetname Benchmark,
    fonttitle=\bfseries\large,
    colframe=black!75!black,
    colback=white!10!white,
    coltitle=white,  
    colbacktitle=gray,  
    boxrule=0.2mm,
    sharp corners,
    shadow={1mm}{-1mm}{0mm}{black!50!white},
    attach boxed title to top left={yshift=-3mm, xshift=3mm},
    boxed title style={sharp corners, size=small}
]

\textbf{Task:} \\
\textsl{[Images]}\\
\texttt{You are very good in question making from documents. I am giving you a task to make some questions from some pages from a document.}

\textbf{Instructions:}
\begin{itemize}
    \item \texttt{The questions should be precise. Each question should be answerable from a very small portion of the document and relevant to the textual and visual elements of the provided image.}
    \item \texttt{Questions should be natural and easy to understand. yet,questions should be challenging enough that even you would find them difficult to answer immediately.}
    \item \texttt{Ensure the questions are open-domain so that even if multiple documents are provided, the question remains uniquely identifiable and answerable.}
    \item \texttt{Include all necessary information to make the question unique and answerable. Avoid vague references like "according to the given article" or "mentioned in the article". Explicitly include the full information if needed.}
    \item \texttt{Create only 2-3 high-quality questions. If a quality question cannot be made, return "NA". However, ensure that effort is made to create a good question.}
    \item \texttt{\textbf{Accepted Questions:}\\
    - "Question": "Who accused AAP of supporting 'terrorist sympathizers' during Punjab elections?"  \\
      "Answer":Anurag Thakur"\\
    - "Question": "What was the altitude of Sandakphu where the tourist died?"\\  
      "Answer":  "11,900 feet"
    }
    \item \texttt{\textbf{Rejected Questions:}\\
    - "Question": "Who is the alleged associate of Partha Chatterjee mentioned in the article?"  \\
      Don't make such questions that reference the artcile.\\
    - "Question": "Which company is prominent in biodiversity monitoring using AI?"\\
      Such question is not acceptable because it is document specific. There can be multiple answers.
    }
    \item \texttt{Stick to the above format. If you are unable to create quality questions, return NA.}

\textbf{Output Format (JSON):}
\begin{verbatim}
{
    "questions": [
        {
            "question": "the question",
            "answer": "the answer"
        },
        ...
    ]
}
\end{verbatim}
\end{itemize}

\end{tcolorbox}






\end{document}